\PassOptionsToPackage{table}{xcolor} 
\documentclass[letterpaper]{article} 
\usepackage[preprint]{aaai2027}
\usepackage[hyphens]{url}  
\usepackage{graphicx} 
\usepackage{natbib}  
\usepackage{caption} 
\usepackage{algorithm}
\usepackage{algorithmic}
\usepackage{multirow}
\usepackage{enumitem}

\usepackage{booktabs}
\usepackage{colortbl}
\usepackage{pifont}
\usepackage{array}
\usepackage{makecell}
\usepackage{tabularx}
\usepackage{bm}
\usepackage{amsmath}
\usepackage{amssymb}
\usepackage{bbding}
\usepackage{threeparttable}
\usepackage{xspace}

\usepackage{hyperref}

\definecolor{headerbg}{gray}{0.9}

\makeatletter
\let\iap@orig@maketitle\@maketitle
\renewcommand{\@maketitle}{%
  \iap@orig@maketitle
  \vspace{0.5em}
    \includegraphics[width=\textwidth]{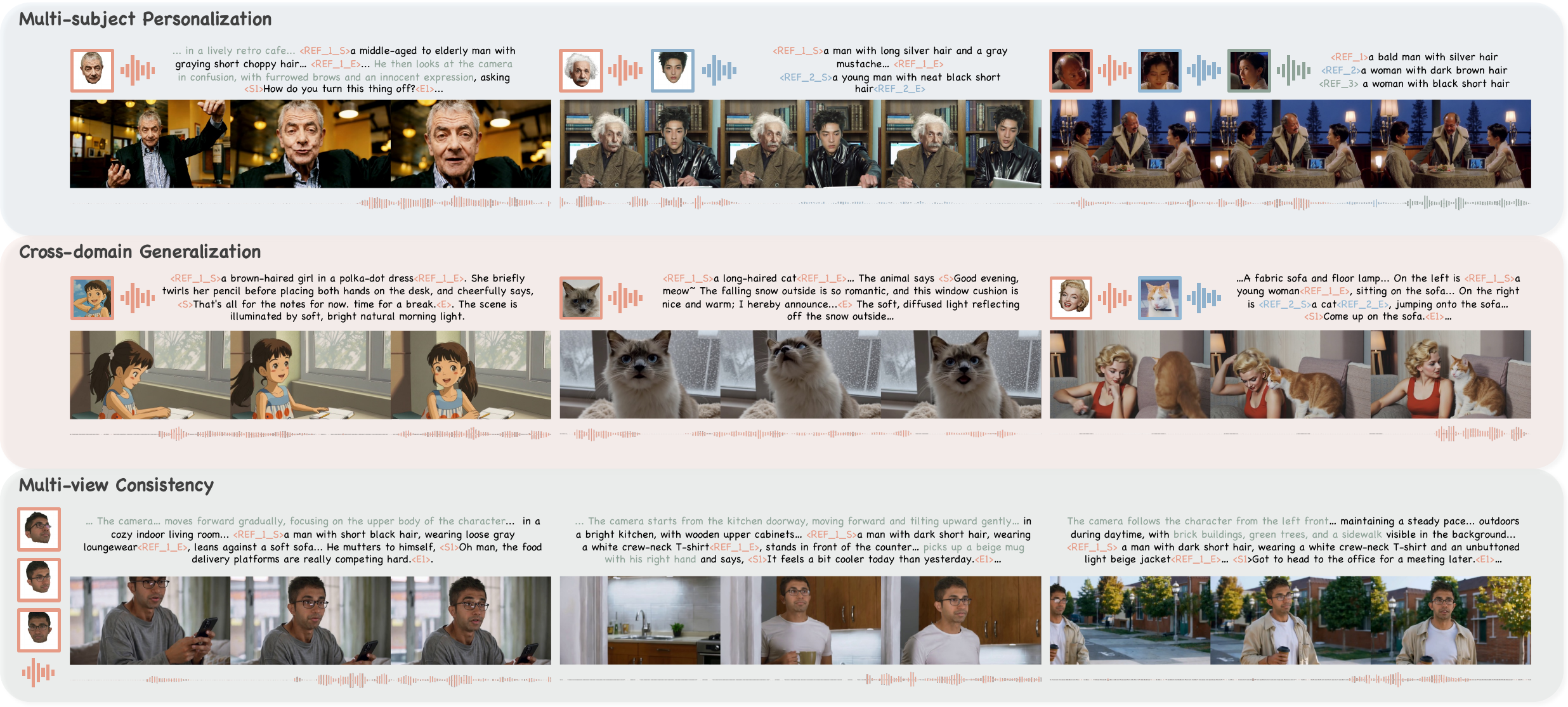}\\[0.35em]
    \refstepcounter{figure}%
    {\small\textbf{Figure~\thefigure.} Identity-as-Presence supports multi-subject personalization, cross-domain personalization, and spatio-temporal multi-view consistent personalization for both appearance and voice.}
    \label{fig:teaser}
  \vspace{1.5em}
}

\newcommand{\tagstart}[1]{\ensuremath{\text{\bfseries\small\ttfamily<#1\_S>}}}
\newcommand{\tagend}[1]{\ensuremath{\text{\bfseries\small\ttfamily<#1\_E>}}}
\newcommand{\speechstart}[1]{\ensuremath{\text{\bfseries\small\ttfamily<S#1>}}}
\newcommand{\speechend}[1]{\ensuremath{\text{\bfseries\small\ttfamily<E#1>}}}

\newcommand{\suppincludegraphics}[2][]{%
  \IfFileExists{#2}{%
    \includegraphics[#1]{#2}%
  }{%
    \centering\fbox{\begin{minipage}[c][0.22\textheight]{0.92\textwidth}\centering
      Missing figure\\{\small\ttfamily\detokenize{#2}}
    \end{minipage}}%
  }%
}

\makeatother

\usepackage{newfloat}
\usepackage{listings}
\DeclareCaptionStyle{ruled}{labelfont=normalfont,labelsep=colon,strut=off} 
\floatstyle{ruled}
\newfloat{listing}{tb}{lst}{}
\floatname{listing}{Listing}

\usepackage{booktabs}
\usepackage{amsmath}

\usepackage{xspace}

\makeatletter
\DeclareRobustCommand\onedot{\futurelet\@let@token\@onedot}
\def\@onedot{\ifx\@let@token.\else.\null\fi\xspace}

\def\ie{\emph{i.e}\onedot}

\makeatother
\title{Identity as Presence: Towards Appearance and Voice Personalized \\ Joint Audio-Video Generation}
\author{
Qin Chen\textsuperscript{\rm 1},
Yingjie Chen\textsuperscript{\rm 1},
Shilun Lin\textsuperscript{\rm 1},
Cai Xing\textsuperscript{\rm 1},
Binxin Yang\textsuperscript{\rm 1},
Long Zhou\textsuperscript{\rm 1},\\
Qixin Yan\textsuperscript{\rm 1},
Wenjing Wang\textsuperscript{\rm 1},
Dingming Liu\textsuperscript{\rm 2},
Hao Liu\textsuperscript{\rm 1},
Chen Li\textsuperscript{\rm 1},
Jing LYU\textsuperscript{\rm 1}
}
\affiliations{
\textsuperscript{\rm 1}WXG, Tencent Inc.\\
\textsuperscript{\rm 2}Peking University\\
\{qinqinchen,alexyjchen,cirolin,yolocai,binxinyang,simonlzhou,qixinyan,augustawang,leweshaoliu,chaselli,eckolv\}@tencent.com,\\
dingmingliu3722@gmail.com
}

\begin{document}

\maketitle

\begin{abstract}
    Recent advances in video synthesis have enabled realistic integration of real individuals, driving demand for identity-aware generation. While emerging methods support joint appearance and voice injection in audio-visual models, they primarily focus on single-subject settings. Multimodal identity integration across multiple subjects remains limited, and precise alignment between visual and vocal identities in multi-subject scenarios remains underexplored.
    We present Identity-as-Presence, a unified framework for joint personalized audio-video generation.
    An automated data curation pipeline constructs identity-labeled audio-visual pairs for single- and multi-subject scenes.
    A unified identity injection mechanism then binds paired appearance and voice through shared cross-modal identity binding and subject-anchored captions.
    A multi-stage training strategy further leverages large-scale unimodal data alongside scarce paired clips to mitigate modality imbalance.
    Experiments show superior audio quality, video fidelity, and audio-visual consistency, with stronger multi-subject binding than the compared methods. For more details and qualitative results, please refer to our webpage: \href{https://chen-yingjie.github.io/projects/Identity-as-Presence}{Identity-as-Presence}.


\end{abstract}

\section{Introduction}

Recent advances in audio-visual generation have established a strong foundation for high-fidelity multimodal synthesis. Building on this progress, emerging applications have created increasing demand for personalized video generation that faithfully replicates both the facial appearance and vocal timbre of real individuals. This trend highlights the need for models with \textit{cameo capabilities}, enabling a specific subject to speak arbitrary content and appear in diverse scenes conditioned on user input. Such functionality is critical for applications including virtual avatars and digital identity preservation, which require precise and disentangled control over facial identity and voice characteristics.

A straightforward solution is a multi-stage cascade. It first generates audio independently and then performs reference-based audio-driven lip synchronization. However, such sequential pipelines impose rigid visual constraints and offer limited support for cinematic dynamics. They also model audio mainly through lip motion rather than holistic sound, which weakens overall audio-visual coherence. These limitations make end-to-end generation a more promising paradigm for unified audio-visual synthesis.

Despite this potential, building a unified identity-personalized joint audio-video model remains difficult.
Unimodal identity control is comparatively mature: audio systems clone timbre from a reference utterance~\cite{ren2019fastspeech,ren2020fastspeech,wang2023neural}, and video systems preserve faces via cross-attention~\cite{ye2023ip,he2024id,yuan2025identity}, parallel branches~\cite{zhang2023adding,hu2023animateanyone}, or token concatenation~\cite{liu2025phantom,fei2025skyreels,xue2025stand}, with recent progress on multi-subject spatial composition~\cite{wang2025interacthuman,deng2025magref,liang2025movie}.
Concurrent works have made progress toward joint audio-visual generation conditioned on both appearance and voice references~\cite{li2026omnicustom,li2026mmcontrol,li2026unitalking,chen2026omni}.
Nevertheless, appearance-voice binding is often insufficiently enforced, and identity entanglement remains prevalent in interactive multi-subject scenarios.
Text prompts likewise provide limited control over \emph{who} should speak or appear when multiple references are given.
Identity-labeled audio-visual pairs also remain scarce relative to unimodal corpora, which complicates stable joint training.

To this end, we present \textbf{Identity-as-Presence}, a unified framework for appearance- and voice-personalized joint audio-video generation (Figure~\ref{fig:teaser}).
An automated curation pipeline scales identity-labeled audio-visual pairs for single- and multi-subject scenes.
On these pairs, shared cross-modal identity binding couples appearance and voice, while subject-anchored captions specify who appears and who speaks under multiple references.
Multi-stage training further mitigates the modality imbalance between abundant unimodal corpora and scarce paired clips.
Experiments show superior audio quality, video fidelity, and audio-visual consistency, with stronger multi-subject binding than the compared state-of-the-art methods.
Our main contributions are threefold.
\begin{itemize}
\item We introduce a data curation pipeline based on detection, diarization, captioning, and filtering that yields identity-labeled audio-visual pairs for multi-subject scenes, where paired supervision is especially scarce.

\item We propose identity injection with shared embeddings, reference positioning, decoupled asymmetric attention, and subject-anchored captions, reducing cross-subject entanglement and appearance-voice mismatch in multi-reference generation.

\item We devise a three-stage curriculum from unimodal pretraining to joint alignment and multi-view fine-tuning, exploiting abundant unimodal data to stabilize training under limited paired clips.
\end{itemize}

\section{Related Work}

\noindent \textbf{Identity-Aware Video Generation.}
Facial identity is commonly injected with face adapters~\cite{he2024id}, temporal consistency constraints~\cite{yuan2025identity}, or reference-token concatenation under full-sequence denoising~\cite{liu2025phantom,fei2025skyreels,jiang2025vace}.
Foundation-model customizers add modality-specific modules at higher cost~\cite{hu2025hunyuancustom,huang2025conceptmaster}, while lightweight designs such as Stand-In~\cite{xue2025stand} reduce trainable parameters.
These methods focus on visual fidelity and do not model appearance-voice binding for identity-personalized joint audio-video generation.

\vspace{1mm}
\noindent \textbf{Identity-Aware Audio Generation.}
Zero-shot speaker cloning covers AR codec language models~\cite{wang2023neural} and efficient NAR or flow-matching TTS~\cite{ren2019fastspeech,ren2020fastspeech,ju2024naturalspeech,mehta2024matcha,eskimez2024e2,chen2025f5}.
Despite strong timbre transfer, these methods lack visual identity and lip-speech coupling.
Many TTS pipelines also rely on audio infilling, which is ill-suited to open-text audio-visual generation.

\vspace{1mm}
\noindent \textbf{Joint Audio-Visual Generation.}
Dual-tower models synthesize video and audio together~\cite{ltx-2,uniavgen,universe,javisdit,low2025ovi}, with symmetric~\cite{low2025ovi,ltx-2} or asymmetric~\cite{javisdit,universe,uniavgen} cross-modal fusion.
Identity-aware extensions personalize single subjects via token concatenation, cross-attention, or bypass branches~\cite{li2026omnicustom}, but offer limited support for disentangling multiple appearance-voice pairs.
For multi-subject control, DreamID-Omni~\cite{guodreamid} and Omni-Customizer~\cite{chen2026omni} rely on specific RoPE settings and attention masks to bind multimodal identities of different subjects.
Such positional or textual cues can fail under crowded prompts or imperfect captions, leaving appearance-voice entanglement unresolved.
We instead bind paired visual and auditory references with shared identity embeddings, use subject-anchored captions for grounding, and apply progressive training to exploit scarce paired audio-visual identity data after unimodal pretraining.

\section{Identity-as-Presence}


We propose Identity-as-Presence, a unified audio-video generation framework for coherent multi-subject control. It consists of three key aspects: automated identity-aware data curation, shared embeddings for unified visual-auditory identity injection, and multi-stage training to mitigate modality disparity and stabilize convergence (Figure~\ref{fig:framework}).

\begin{figure*}[t]
    \centering
    \includegraphics[width=0.95\textwidth]{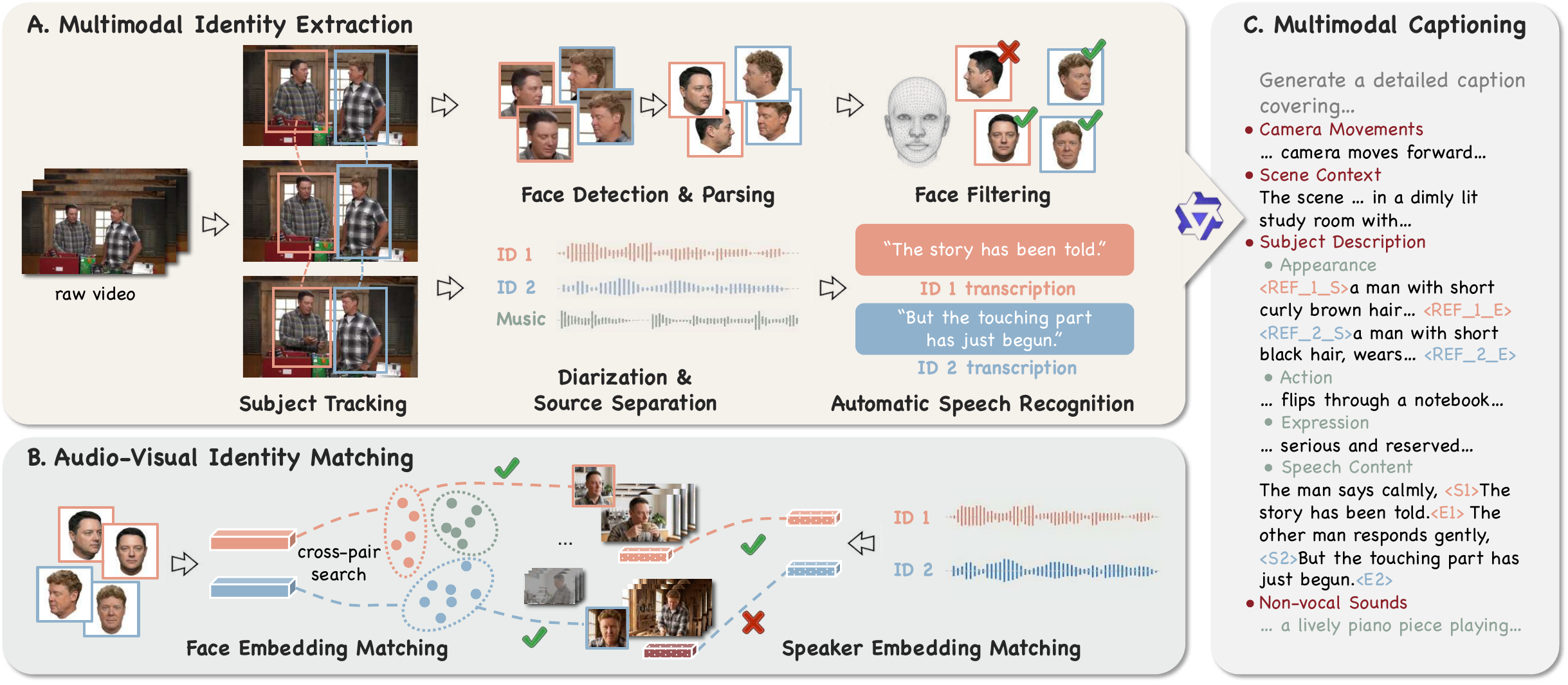}
    \caption{Overview of the data curation pipeline: A. extracting paired appearance-voice identity signals from raw videos, B. matching cross-clip references for identity injection, and C. synthesizing structured captions with subject anchors.}
    \label{fig:data_curation}
\end{figure*}

\subsection{Preliminaries}

To address the distinct dynamics of visual and auditory modalities, a dual-tower architecture is employed, facilitating cross-modal interaction via attention mechanisms.
Let $v_0$ and $a_0$ denote the clean latents encoded by their respective VAEs. Independent probability paths are defined via linear interpolation with Gaussian noise $v_1, a_1 \sim \mathcal{N}(0, I)$:
\begin{equation}
    v_t = (1 - t)v_0 + t v_1, \quad a_t = (1 - t)a_0 + t a_1,
\end{equation}
where $t \in [0, 1]$.
These paths induce target velocity fields $u_t^v = v_1 - v_0$ and $u_t^a = a_1 - a_0$. To enable personalized generation, the model is conditioned on disentangled visual and auditory identity signals $c^v$ and $c^a$. Consequently, the joint optimization objective is a weighted sum of two flow matching losses:
\begin{equation}
\begin{aligned}
\mathcal{L}_{\text{joint}}
= \mathbb{E}_{t,v_0,a_0,v_1,a_1}\Big[
&\left\|\hat{u}_{\theta}^{v}(v_t,a_t,t,c^v)-u_t^v\right\|_2^2 \\
&+ \lambda
\left\|\hat{u}_{\theta}^{a}(v_t,a_t,t,c^a)-u_t^a\right\|_2^2
\Big].
\end{aligned}
\end{equation}
where $\hat{u}_{\theta}^v$ and $\hat{u}_{\theta}^a$ denote the predicted velocity fields conditioned on their respective identity controls, and $\lambda$ balances the learning between the two modalities.

\subsection{Data Curation Pipeline}
\label{sec:data_curation}

Paired identity-labeled audio-visual data are scarce but essential for learning appearance-voice association.
We therefore build an automated curation pipeline (Figure~\ref{fig:data_curation}) that converts raw videos from OpenHumanVid~\cite{li2025openhumanvid} and OmniHuman~\cite{zhu2026omnihuman} into training tuples $\mathcal{D}=\{(v,a,c^v,c^a,\mathrm{txt})\}$ for single- and multi-subject scenes.
Beyond scale, the pipeline enforces correct appearance-voice pairing and non-trivial reference-target gaps, so the model must inject identity rather than copy content.
More details and statistics are provided in the supplementary material.

\vspace{1mm}
\noindent\textbf{Multimodal Identity Extraction.}
We detect humans with YOLOv11~\cite{khanam2024yolov11}, track them with MOTRv2~\cite{zhang2023motrv2}, and estimate FLAME~\cite{flame} parameters with SMIRK~\cite{retsinas2024smirk} to obtain one-shot or spatio-temporal multi-view appearance references $c^v$.
On the audio side, Demucs~\cite{defossez2019demucs} separates speech, 3D-Speaker~\cite{chen20253d} performs diarization, and ASR yields per-subject transcripts for voice references $c^a$.
SyncNet~\cite{chung2016out} then retains only lip-sync-consistent appearance-voice pairs.

\vspace{1mm}
\noindent\textbf{Multimodal Captioning with Subject Anchors.}
Given the clip, face crops, and transcripts, Qwen3-Omni~\cite{qwen-omni} generates structured captions of camera, scene, ambience, actions, and speech.
We require \emph{subject anchors} that bind appearance and spoken content to unique subject IDs, enabling text to specify who appears and who speaks.
The full prompt is provided in the supplementary material.

\vspace{1mm}
\noindent\textbf{Cross-clip Audio-visual Identity Matching.}
To discourage copy-and-paste, we match references across clips of the same subject with varied pose/expression and non-overlapping speech content.
Clips are first grouped by ArcFace~\cite{deng2019arcface} embeddings and then verified with ERes2Net~\cite{chen2024eres2netv2} speaker embeddings to reject dubbing and voice-over mismatches.

The resulting supervision supports identity-aware joint audio-video learning, including multi-subject scenes that require unambiguous appearance-voice association.

\subsection{Network Architecture}

\begin{figure*}[t]
    \centering
    \includegraphics[width=0.95\textwidth]
    {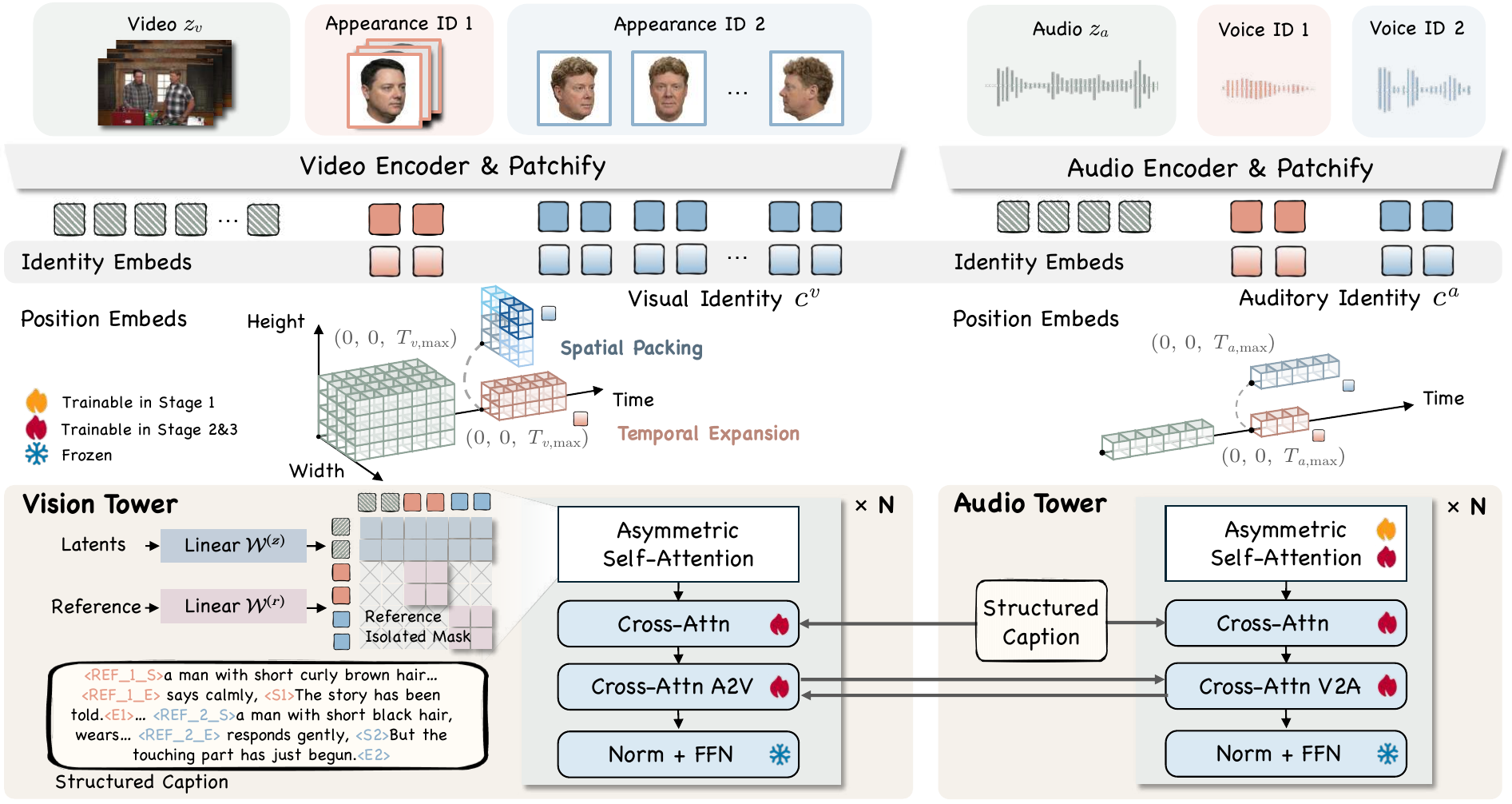}
     \caption{The overall dual-tower DiT architecture and training framework. The model encodes video, audio, paired appearance/voice references, and subject-anchored captions, binds them with shared identity embeddings, and injects them via reference positioning and decoupled asymmetric attention. Training proceeds in three stages: unimodal identity, joint multimodal alignment, and spatio-temporal multi-view fine-tuning.}
    \label{fig:framework}
\end{figure*}

To achieve high-fidelity joint audio-video generation, we build on a dual-tower Diffusion Transformer (DiT). The key challenge is to inject facial appearance and vocal timbre in a disentangled yet coherent way. We address this with a unified identity injection mechanism that combines shared identity embeddings, reference positioning, and decoupled asymmetric attention, as shown in Figure~\ref{fig:framework}.

\subsubsection{Multimodal Identity Injection.}
Instead of employing auxiliary control networks or adapters, we treat identity signals as additional input tokens and seamlessly integrate them into the input sequence of the DiT backbone.
This unified approach allows for the joint optimization of multimodal content generation and identity preservation.

Let $m \in \{v, a\}$ denote the modality index. Let $z_t^m \in \mathbb{R}^{L_t^m \times D}$ be the sequence of noisy latent tokens at diffusion timestep $t$, and $c^m$ be the corresponding raw reference signal. Specifically, for the visual branch ($m=v$), $c^v$ comprises reference facial images (supporting one-shot as well as spatio-temporal multi-view layouts), and for the auditory branch ($m=a$), $c^a$ consists of a reference audio clip. We first encode these control signals into discrete tokens $r^m$ using modality-specific VAE encoders:
\begin{equation}
    r^m = \text{Patchify}(\mathcal{E}_m(c^m)) \in \mathbb{R}^{L_r^m \times D}.
\end{equation}

\vspace{1mm}
\noindent \textbf{Shared Identity Embeddings.}
To enable robust multi-subject personalized generation, we must ensure that each voice timbre is correctly associated with its corresponding facial appearance.
To this end, we introduce learnable \textit{Identity Embeddings} to explicitly bind these modalities at the token level.
For a scene with $n$ distinct identities ($n\le K_{\max}$), we assign each subject a slot index $k$ and the corresponding vector $e^{id}_{(k)}$. This embedding is injected into the reference tokens via element-wise addition:
\begin{equation}
        \hat{r}^m_{(k)} = r^m_{(k)} + e^{id}_{(k)}, \quad m \in \{v, a\}.
\end{equation}
Since $e^{id}_{(k)}$ is shared between $r^v_{(k)}$ and $r^a_{(k)}$, it serves as an explicit binding signal that guides the model to associate the visual tokens of identity $k$ with the auditory tokens of the same identity.
In training, we set a fixed number of identity slots, and subject-to-slot assignments are randomly permuted each iteration so that every slot receives a gradient under the long-tailed subject-count distribution.
Identity-bound reference tokens $\hat{r}^m$ are then concatenated with noisy latents $z_t^m$ along the sequence dimension to form $X^m \in \mathbb{R}^{(L_r^m + L_t^m) \times D}$ as the composite input for each tower:
\begin{equation}
    X^m = [\hat{r}^m; z_t^m].
\end{equation}
Token-wise concatenation introduces a symmetric inductive bias that enables the model to flexibly attend to fine-grained features, facilitating the aggregation of spatio-temporal, multi-view facial cues and the extraction of spectral formant representations with minimal architectural overhead.

\vspace{1mm}
\noindent \textbf{Reference Position Embeddings.}
Appearance-voice binding and cross-subject distinction are carried by the shared identity embeddings $e^{id}_{(k)}$. RoPE only encodes spatio-temporal geometry for generation and reference layout, not identity.
For noisy latents $z_t^v$ and $z_t^a$, factorized 3D RoPE~\cite{su2021roformer} is applied for video and scaled 1D RoPE for audio.
For reference tokens, we place all identities at a shared margin beyond the generation horizon,
\ie, every visual reference starts at $T_{v,\max}$, and every auditory reference starts at $T_{a,\max}$ (audio references occupy a contiguous span of length up to $L$).
Placing all references beyond $T_{\cdot,\max}$ avoids occupying the early temporal indices and interfering with frame-conditioned generation.
To further improve view diversity and dynamic consistency of each subject, we use two complementary layouts that target different multi-view evidence:
\textbf{(i)~spatial packing} tiles multi-view images lacking temporal dependencies in a near-square grid within $(H,W)$ at the shared margin index $T_{v,\max}$ before the patchify operation, favoring \emph{multi-view consistency} under large pose or camera changes;
\textbf{(ii)~temporal expansion} places images of the same subject along consecutive indices beyond $T_{v,\max}$ with shared spatial anchors, favoring \emph{dynamic consistency} (e.g., expression transitions) over a short temporal process.

\vspace{1mm}
\noindent \textbf{Decoupled Asymmetric Attention.}
Standard self-attention layers typically share projection weights across all tokens. However, given the distinct semantic roles of the reference conditions and the noisy latents, sharing weights may lead to optimization conflicts. To this end, we use decoupled projection weights together with an asymmetric attention mask.

Let $X \in \mathbb{R}^{L \times D}$ be the input sequence. We partition token indices into a noisy latent set $\Omega_z$ and a reference set $\Omega_r=\bigcup_{k\in\mathcal{S}}\Omega_r^{(k)}$, where $\mathcal{S}$ is the set of occupied identity slots in the clip and $\Omega_r^{(k)}$ collects all reference tokens assigned to slot~$k$ (visual and auditory). Thus $\Omega_r\cup\Omega_z=\{1,\dots,L\}$.
To learn specialized representations, we use separate projection weights for the two groups: $\mathcal{W}^{(r)}$ for reference tokens and $\mathcal{W}^{(z)}$ for noisy tokens. Each token is projected using the weights corresponding to its set, producing the matrices $Q, K, V \in \mathbb{R}^{L \times d}$. This decoupling allows the model to learn identity-preserving and denoising features independently.

To control information flow under multi-reference conditions and prevent interference, we introduce a structural reference isolation mask $M$ for attention computation:
\begin{equation}
\begin{aligned}
\operatorname{Attn}(Q,K,V)
&=
\operatorname{softmax}
\left(
\frac{QK^\top}{\sqrt{d}} + M
\right)V, \\
M_{i,j}
&=
\begin{cases}
-\infty, & i \in \Omega_r^{(k)},\; j \in \Omega_z, \\
-\infty, & i \in \Omega_r^{(k)},\; j \in \Omega_r^{(\ell)},\; k \neq \ell, \\
0,       & \text{otherwise.}
\end{cases}
\end{aligned}
\end{equation}
With this mask, the main latents can extract identity cues from all references, whereas each reference remains isolated from both diffusion noise and other references, preserving the integrity of the control signals during generation.

\setlength\intextsep{-10pt}
\begin{table}[t]
\centering
\resizebox{\columnwidth}{!}{%
\footnotesize
\setlength{\tabcolsep}{1.2pt}
\renewcommand{\arraystretch}{1.02}
\begin{tabular}{l|cccc}
\toprule
    \multirow{2.5}{*}{Method} 
    & \multicolumn{2}{c}{English} 
    & \multicolumn{2}{c}{Chinese} \\
    \cmidrule(lr){2-3} \cmidrule(lr){4-5} 
    & WER\%$\downarrow$ 
    & AID-SIM$\uparrow$ 
    & WER\%$\downarrow$ 
    & AID-SIM$\uparrow$ \\
\midrule
E2 TTS~\cite{eskimez2024e2} & 2.19 & 0.710 & 1.97 & 0.730 \\
F5 TTS~\cite{chen2025f5} & \underline{1.83} & 0.647 & 1.56 & 0.741 \\
\midrule
CosyVoice2~\cite{du2024cosyvoice} & 2.57 & 0.652 & 1.45 & 0.748 \\
CosyVoice3~\cite{du2025cosyvoice} & 2.21 & \underline{0.720} & \textbf{1.12} & \underline{0.781} \\
\midrule
\textbf{Ours-LTX-2.3 Audio} & \textbf{1.57} & \textbf{0.770} & \underline{1.20} & \textbf{0.785} \\
\bottomrule
\end{tabular}%
}
\vspace{-2mm}
\caption{Comparison with SOTA TTS Methods on Seed-TTS-Eval~\cite{anastassiou2024seed}.}
\label{tab:comp_tts}
\vspace{-2mm}
\end{table}

\begin{table*}[!htt]
\centering
\resizebox{\textwidth}{!}{%
\footnotesize
\setlength{\tabcolsep}{1.2pt}
\renewcommand{\arraystretch}{1.02}
\begin{tabular}{@{}l@{}|ccccc|cccc|ccc|cc@{}}
    \toprule
    \multirow{2}{*}{\textbf{Models}} 
    & \multicolumn{5}{c|}{\textbf{Audio}} 
    & \multicolumn{4}{c|}{\textbf{Video}} 
    & \multicolumn{3}{c|}{\textbf{A/V Consistency}} 
    & \multicolumn{2}{c}{\textbf{Binding$^\dagger$}} \\
    \cmidrule(l{2pt}r{2pt}){2-6} \cmidrule(l{2pt}r{2pt}){7-10} \cmidrule(l{2pt}r{2pt}){11-13} \cmidrule(l{2pt}r{2pt}){14-15}
    &
    PQ$\uparrow$
    & CLAP$\uparrow$
    & FD$\downarrow$ 
    & WER$\downarrow$ 
    & AID-SIM$\uparrow$
    & AES$\uparrow$ 
    & DD$\uparrow$ 
    & OC$\uparrow$
    & VID-SIM$\uparrow$ 
    & Sync-C$\uparrow$ 
    & Sync-D$\downarrow$ 
    & ImageBind$\uparrow$
    & FM$\uparrow$
    & SM$\uparrow$ \\ 
    \midrule
    Phantom$^{\ddagger}$
      & / & / & / & / & / & 0.54/0.53 & 0.55/0.57 & \underline{0.11}/0.09 & 0.74/0.49 & / & / & / & 0.62 & / \\
    \midrule
    CV3+HunyuanCustom~~
      & \underline{7.32} & 0.14 & 1.00 & \textbf{22.94} & 0.48 & 0.54 & 0.60 & \underline{0.11} & 0.64 & 5.41 & 8.66 & 0.27 & / & / \\
    CV3+HuMo
      & \textbf{7.35} & 0.10 & 1.06 & \underline{24.09} & 0.47 & 0.54 & 0.33 & \underline{0.11} & 0.71 & 5.35 & 8.86 & 0.28 & / & / \\
    \midrule
    QIE+UniAVGen
      & 6.15 & \underline{0.24} & 0.97 & 68.86 & 0.32 & 0.35 & 0.65 & \underline{0.11} & 0.37 & 3.99 & 11.66 & 0.29 & / & / \\
    QIE+ID-LoRA
      & 7.03 & 0.11 & 0.84 & 30.59 & \underline{0.54} & 0.56 & 0.58 & \underline{0.11} & 0.41 & 6.30 & 7.85 & \underline{0.36} & / & / \\
    OmniCustom
      & 6.10 & 0.13 & 0.90 & 33.71 & 0.34 & 0.57 & 0.14 & 0.10 & 0.35 & 6.45 & 8.41 & 0.22 & / & / \\
    DreamID-Omni
      & 5.97/5.63 & 0.22/\underline{0.23} & 0.91/0.78 & 28.48/39.99 & 0.39/0.30 & 0.56/0.54 & 0.42/0.40 & \underline{0.11}/\underline{0.10} & 0.67/0.42 & 4.84/4.02 & 9.48/10.80 & 0.25/0.22 & 0.57 & 0.62 \\
    Ours-Ovi
      & 6.52/\underline{6.15} & 0.19/0.20 & \textbf{0.76}/\underline{0.75} & 32.89/\underline{28.10} & \underline{0.54}/\textbf{0.47} & \underline{0.58}/\textbf{0.57} & \underline{0.69}/\underline{0.64} & \underline{0.11}/\underline{0.10} & \underline{0.75}/\underline{0.59} & \textbf{6.76}/\underline{4.63} & \underline{7.63}/\textbf{9.57} & \textbf{0.38}/\textbf{0.33} & \underline{0.70} & \underline{0.67} \\
    Ours-LTX-2.3
      & 7.30/\textbf{6.24} & \textbf{0.29}/\textbf{0.25} & \underline{0.78}/\textbf{0.72} & 24.77/\textbf{27.86} & \textbf{0.57}/\underline{0.45} & \textbf{0.59}/\underline{0.55} & \textbf{0.70}/\textbf{0.67} & \textbf{0.12}/\textbf{0.11} & \textbf{0.78}/\textbf{0.60} & \underline{6.74}/\textbf{5.05} & \textbf{7.44}/\underline{10.37} & \textbf{0.38}/\underline{0.30} & \textbf{0.85} & \textbf{0.70} \\
    \bottomrule
    \multicolumn{15}{l}{\scriptsize $^\dagger$FM/SM on multi-subject only (R2AV methods with multi-identity support). $^{\ddagger}$No generated audio. CV3: CosyVoice3; QIE: Qwen-Image-Edit. Entries are single/multi when both splits are available.}
\end{tabular}%
}
\vspace{-3mm}
\caption{Comparison under a matched identity-conditioned protocol. Bold and underline mark best and second-best over all reported scores for each metric (single and multi ranked separately).}
\label{tab:comp_sota}
\vspace{-2mm}
\end{table*}

\begin{figure*}[!t]
    \centering
    \includegraphics[width=0.98\textwidth]
    {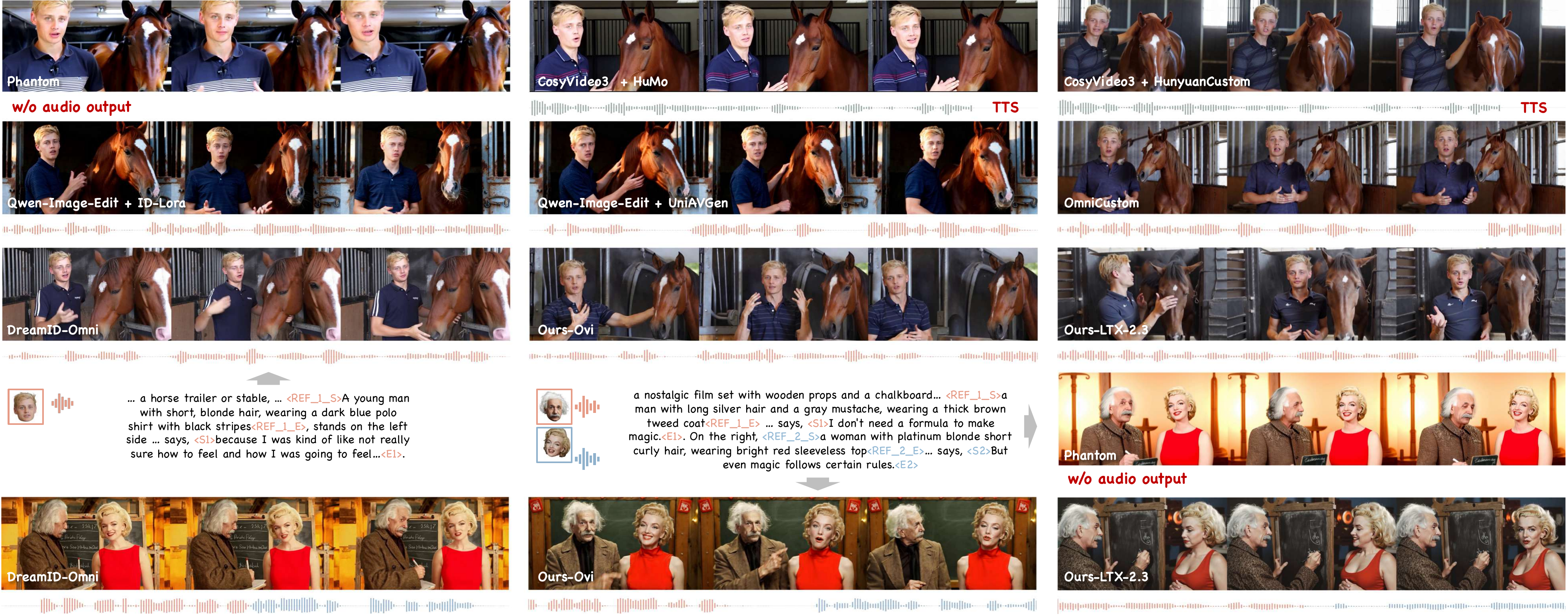}
    \vspace{-3mm}
    \caption{Qualitative comparison with identity-aware video and joint audio-video baselines.}
    \label{fig:comparison}
    \vspace{-2mm}
\end{figure*}

\subsection{Training Strategy}
\label{sec:training}

To leverage heterogeneous data across modalities and facilitate progressive audio-visual alignment, we adopt a multi-stage training strategy motivated by disparities in data availability.
Identity-labeled audio data (e.g., TTS corpora) is the most abundant, identity-labeled video data is more limited due to filtering requirements, and paired identity-labeled audio-visual data is the scarcest because of strict synchronization and semantic constraints. Directly training on limited paired data risks overfitting and poor identity diversity. We therefore propose a multi-stage training strategy (a curriculum) with three stages: unimodal identity training, joint multimodal identity training, and spatio-temporal multi-view identity fine-tuning.

\vspace{1mm}
\noindent \textbf{Stage 1: Unimodal Identity Training.}
We first train each tower on large unimodal data to learn identity priors before cross-modal alignment.
The audio tower is trained on TTS corpora for timbre and prosody. We find that multilingual speech ability is best acquired at this stage, since large multilingual TTS data can be used without paired video and remains largely preserved after Stages~2-3.
Table~\ref{tab:comp_tts} shows that the audio tower overall surpasses the state-of-the-art TTS models.
In parallel, the video tower is trained on identity-labeled video data with a dummy audio counterpart.


\vspace{1mm}
\noindent \textbf{Stage 2: Joint Multimodal Identity Training.}
Then, we perform joint multimodal training using curated paired identity-labeled audio-visual data. The model is initialized from Stage~1 and cross-modal fusion layers are activated to learn temporal alignment between speech and facial motion.
This decoupling allows large unimodal datasets to provide identity diversity while a smaller paired set enables precise audio-visual synchronization.

\vspace{1mm}
\noindent \textbf{Stage 3: Spatio-temporal Multi-view Fine-tuning.}
We then fine-tune on a small multi-view identity set with the two complementary reference-positioning layouts above (spatial packing and temporal expansion).
Spatial packing is used when references emphasize multi-view consistency under pose/viewpoint change.
Temporal expansion is used when references emphasize dynamic consistency such as expression evolution.
Stage~3 therefore adapts the model to both multi-view types rather than selecting a single winner.

\begin{figure}[!t]
    \centering
    \includegraphics[width=\columnwidth]
    {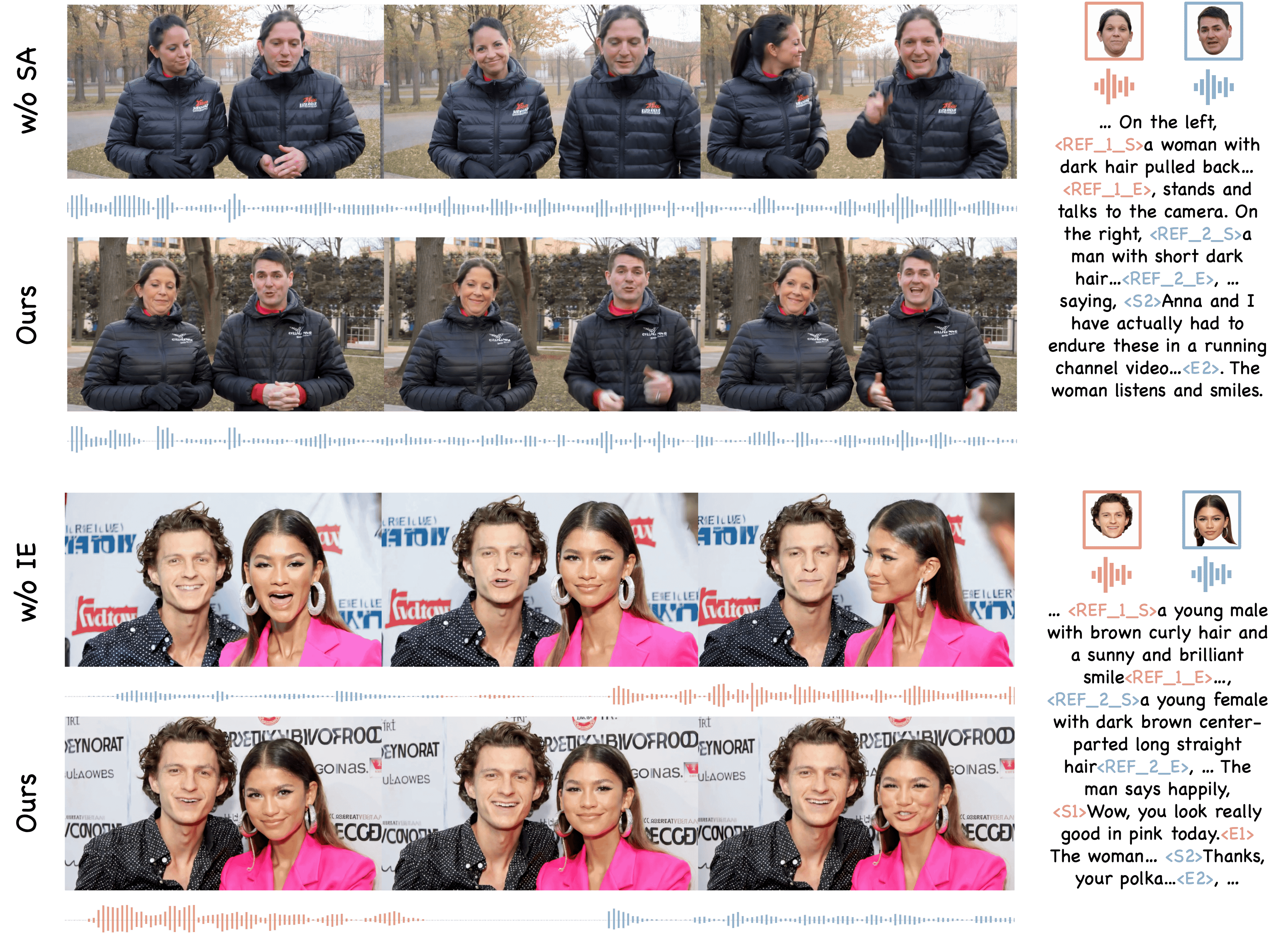}
    \vspace{-0.5em}
    \caption{Typical multi-subject binding failures motivating FM/SM: identity collapse (top), appearance-voice mismatch and overlapped dual talking (bottom).}
    \label{fig:ablation}
    \vspace{-0.5em}
\end{figure}

\section{Experiments}
\label{sec:experiments}

\subsection{Experimental Settings}

We evaluate on an identity-disjoint bilingual benchmark of $400$ triplets ($200$ single-subject/$200$ multi-subject), each with appearance and voice references plus a prompt with bilingual dialogue.
Both splits include challenging cases such as extreme poses, occlusions, complex multi-subject interactions, offscreen subjects, and late-entry subjects.
All methods share matched identity references, and prompts are adapted for each method. For methods that require a reference-based first frame, we use Qwen-Image-Edit~\cite{wu2025qwen} to generate an image from the same references and adapted prompts. 
The supplementary material provides more details.

For audio, we report WER (intelligibility), CLAP (audio-text semantics), FD (distributional quality), PQ (production quality), and AID-SIM (speaker similarity).
For video, we report AES (aesthetic quality) / DD (dynamic degree) / OC (overall consistency) from VBench and VID-SIM (face identity similarity).
For audio-visual consistency, we report Sync-C/Sync-D (lip-sync) and ImageBind (visual and auditory event-level alignment).
Because average AID-SIM / VID-SIM can remain high under cross-binding, we further report Face Match (FM) and Speaker Match (SM) as assignment accuracies on multi-subject cases only (details in Supp.).

We implement Identity-as-Presence on LTX-2.3~\cite{ltx-2} as the primary dual-tower backbone (\textbf{Ours-LTX-2.3}).
Reference audio length is $L{=}5$s and $T_{\max}{=}12$s.
Training follows the three-stage strategy
with unimodal audio/video budgets of $10$k/$10$k iterations and $15$k joint steps, followed by spatio-temporal multi-view fine-tuning.
Optimizer settings, inference hyperparameters, and an additional Ovi~\cite{low2025ovi}-based implementation reported for reference (Ours-Ovi) can be found in the supplementary material (Supp.).

\begin{table}[!t]
\centering
\resizebox{\columnwidth}{!}{%
\footnotesize
\setlength{\tabcolsep}{1.2pt}
\renewcommand{\arraystretch}{1.02}
\begin{tabular}{l|cccc|cc}
    \toprule
    \textbf{Variant}
    & WER$\downarrow$
    & AID-SIM$\uparrow$
    & VID-SIM$\uparrow$
    & Sync-C$\uparrow$
    & FM$\uparrow$
    & SM$\uparrow$ \\
    \midrule
    A~Shared IE
      & \underline{25.47}/\underline{38.91} & \underline{0.56}/\textbf{0.47} & \underline{0.70}/\underline{0.54} & \underline{6.48}/\textbf{5.06} & \underline{0.79} & \underline{0.65} \\
    B~RoPE Offset
      & 28.76/49.36 & 0.44/0.31 & 0.67/0.23 & 5.79/4.48 & 0.58 & 0.59 \\
    C~Rotary Bias
      & 27.43/44.67 & 0.52/0.41 & 0.63/0.21 & 6.20/4.30 & 0.43 & 0.33 \\
    \midrule
    A+Subject Anchors
      & \textbf{24.77}/\textbf{33.86} & \textbf{0.57}/\underline{0.45} & \textbf{0.75}/\textbf{0.59} & \textbf{6.74}/\underline{5.05} & \textbf{0.85} & \textbf{0.70} \\
    \bottomrule
\end{tabular}%
}
\caption{Ablation of binding and caption design. Bold and underline mark best and second-best for each metric.}
\label{tab:ablation_design}
\end{table}

\begin{table}[!t]
\centering
\resizebox{\columnwidth}{!}{%
\footnotesize
\setlength{\tabcolsep}{1.2pt}
\renewcommand{\arraystretch}{1.02}
\begin{tabular}{l|cccc|cc}
    \toprule
    \textbf{Schedule}
    & WER$\downarrow$
    & AID-SIM$\uparrow$
    & VID-SIM$\uparrow$
    & Sync-C$\uparrow$
    & FM$\uparrow$
    & SM$\uparrow$ \\
    \midrule
    One-stage
      & 36.43/51.03 & 0.44/0.31 & 0.61/\underline{0.60} & 5.55/4.26 & 0.65 & 0.51 \\
    Two-stage
      & \underline{24.77}/\underline{33.86} & \underline{0.57}/\underline{0.45} & \underline{0.75}/0.59 & \underline{6.74}/\underline{5.05} & \underline{0.85} & \underline{0.70} \\
    +~ST multi-view FT
      & \textbf{22.39}/\textbf{32.58} & \textbf{0.58}/\textbf{0.54} & \textbf{0.78}/\textbf{0.61} & \textbf{6.80}/\textbf{5.32} & \textbf{0.87} & \textbf{0.71} \\
    \bottomrule
\end{tabular}%
}
\caption{Ablation of training strategy. Bold and underline mark best and second-best for each metric.}
\label{tab:ablation_train}
\end{table}

\begin{figure*}[t]
    \centering
    \includegraphics[width=0.99\textwidth]
    {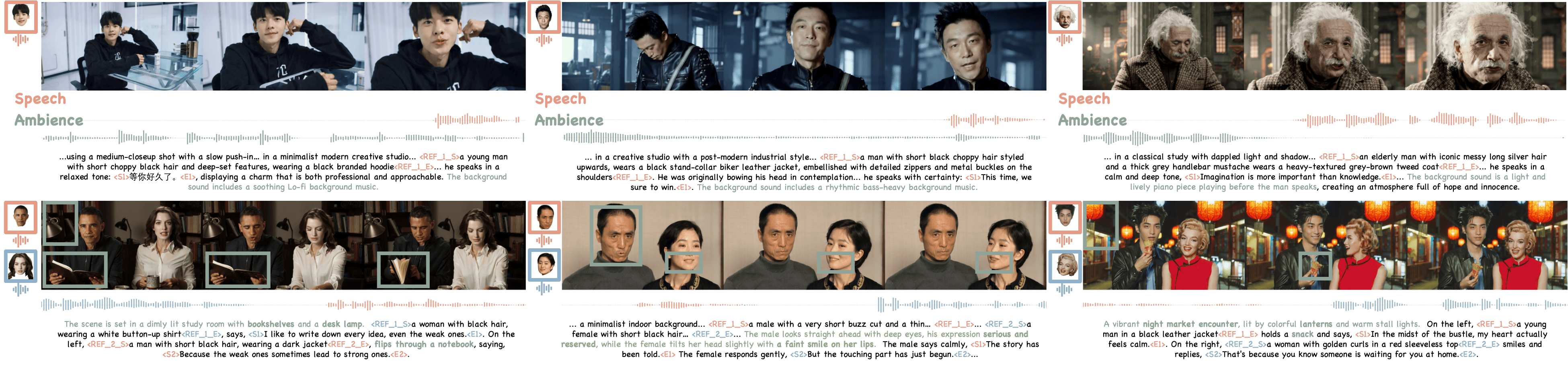}
    \caption{Representative qualitative results for identity-preserving joint audio-video generation.}
    \label{fig:case_study1}
\end{figure*}

\begin{figure}[!t]
    \centering
    \includegraphics[width=0.92\columnwidth]
    {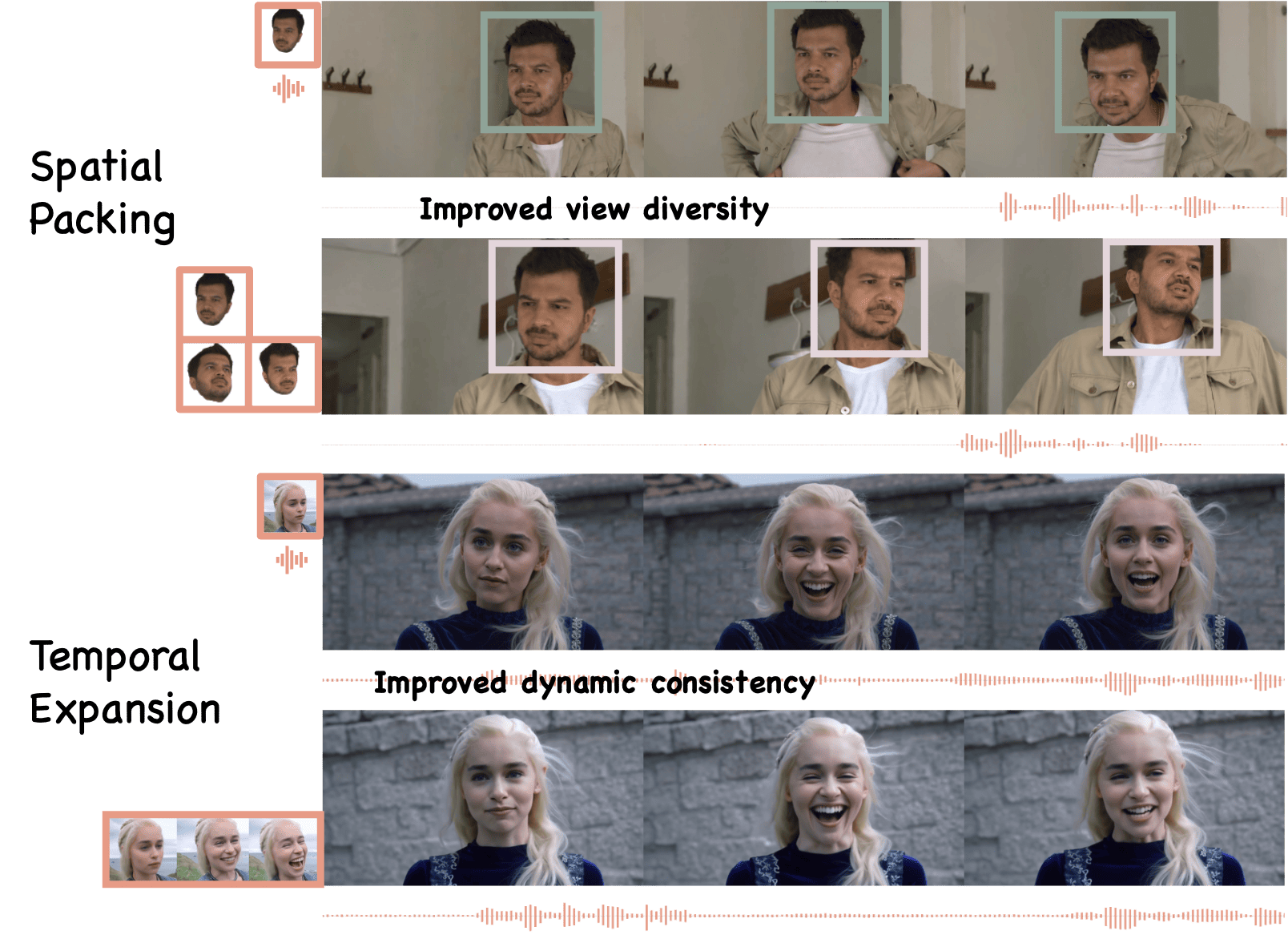}
    \caption{Spatio-temporal multi-view reference positioning.}
    \label{fig:case_study2} 
\end{figure}

\subsection{Comparison with State-of-the-Art Methods}

We compare methods in Table~\ref{tab:comp_sota}, including methods under reference-to-video (R2V), reference-audio-to-video (RA2V, TTS+lip-sync), and reference-to-audio-video (R2AV) protocols (refer to Supp. for details).
All samples are generated at 720p for 5 seconds.
Baselines include R2V Phantom~\cite{liu2025phantom}, RA2V CosyVoice3~\cite{du2025cosyvoice}+HunyuanCustom~\cite{hu2025hunyuancustom}/HuMo~\cite{chen2026human}, and R2AV UniAVGen~\cite{uniavgen}, Qwen-Image-Edit~\cite{wu2025qwen}+ID-LoRA~\cite{dahan2026id}, OmniCustom~\cite{li2026omnicustom}, DreamID-Omni~\cite{guodreamid}, together with our LTX-2.3 implementation \textbf{Ours-LTX-2.3}~\cite{ltx-2}.
We also report Ours-Ovi~\cite{low2025ovi} under the same identity injection mechanism to demonstrate the generality of our method (implementation details in Supp.).
We attribute the lower WER of RA2V to CosyVoice3~\cite{du2025cosyvoice}, a state-of-the-art TTS approach conditioned on ground-truth transcripts. As such, their slightly higher WER primarily reflects the oracle TTS system's high intelligibility, rather than superior performance in identity-personalized joint audio-visual generation.
Our R2AV models are strongest on joint audio-visual consistency and multi-subject binding (FM/SM); \textbf{Ours-LTX-2.3} is better on video fidelity and improves FM/SM over DreamID-Omni under the same assignment protocol.
Relative to R2AV baselines without explicit multi-subject binding, shared identity embeddings reduce cross-talk.
Figure~\ref{fig:comparison} shows qualitative comparisons.

\subsection{Ablation Study}

We ablate binding, captions, and training under matched budgets on the LTX-2.3 implementation (Tables~\ref{tab:ablation_design} and~\ref{tab:ablation_train}).
Dual-identity generation must preserve each appearance and voice, and their pairing.
Figure~\ref{fig:ablation} shows that without SA, multiple subjects collapse onto a shared face, and without shared IE, the model fails to align face and voice IDs and produces voice-face mismatch.
Table~\ref{tab:ablation_design} asks whether shared identity embeddings outperform RoPE-layout binding under the same reference-positioning budget.
\textbf{A} uses shared identity embeddings (IE) with a shared $T_{\cdot,\max}$ margin.
\textbf{B} removes shared IE and encodes each subject by a distinct RoPE time offset (per-subject RoPE offset similar to Syn-RoPE~\cite{guodreamid}).
\textbf{C} adds a per-subject rotary bias to align same-subject cross-modal tokens.
\textbf{A} is most reliable (FM~$0.79$ vs.\ $0.58$/$0.43$ for B/C) and does not entangle identity with temporal offsets as subject count grows.
Adding subject anchors (SA) further improves multi-subject grounding (\textbf{A+SA}).
For scarce paired audio-visual data, Table~\ref{tab:ablation_train} shows the three-stage training strategy outperforms one-stage joint training. The final spatio-temporal multi-view fine-tuning stage further improves identity metrics by exposing both complementary multi-view layouts (refer to Supp. for more details).

\subsection{Qualitative Analysis}
Figure~\ref{fig:case_study1} illustrates faithful prompt following in joint generation.
The top row shows single-subject results that realize audio-related descriptions in the text prompt, including clear speech and realistic non-vocal sounds fused with the visuals.
The second row shows dual-subject cases that respond to fine-grained captions specifying who speaks and who appears.
Despite training only on real-human video, Figure~\ref{fig:teaser} demonstrates zero-shot generalization to cartoon and animal domains not covered by the training data.
Figure~\ref{fig:case_study2} shows the two complementary multi-view layouts. \emph{Spatial packing} better preserves identity under large pose or viewpoint change, while \emph{temporal expansion} better captures dynamic appearance cues such as expression transitions (refer to Supp. for additional qualitative results).

\section{Conclusion}

We presented Identity-as-Presence, a unified framework for appearance- and voice-personalized joint audio-video generation.
Shared identity embeddings provide more reliable appearance-voice association than RoPE positional binding. Subject-anchored captions improve multi-subject text grounding. Progressive training better utilizes scarce paired audio-visual identity data.
Experiments show superior audio quality, video fidelity, and cross-modal alignment, with improved multi-subject binding over the compared methods.
Limitations and failure cases are discussed in Supp.


\bibliography{aaai2027}

\clearpage
\twocolumn[
  \begin{center}
    {\LARGE\bfseries Supplementary Material\par}
    \vspace{1.2em}
  \end{center}
]

\setcounter{figure}{0}
\renewcommand{\thefigure}{\Alph{figure}}
\setcounter{algorithm}{0}
\renewcommand{\thealgorithm}{\Alph{algorithm}}

\section{Data Curation Details}

\subsection{Pipeline modules}
As described in the main paper, we convert raw videos into identity-labeled training tuples $\{(v,a,c^v,c^a,\mathrm{txt})\}$ with our data curation pipeline.
Below we expand the pipeline modules and the role of each component.

\noindent\textbf{Multimodal Identity Extraction.}
From each current clip, we extract per-subject appearance and voice signals and verify that they are lip-sync consistent within the clip.
On the visual side, we construct per-subject tracks and then extract face crops.
YOLOv11~\cite{khanam2024yolov11} detects humans in each frame, and MOTRv2~\cite{zhang2023motrv2} links detections across frames into subject tracks.
For each track, SMIRK~\cite{retsinas2024smirk} regresses FLAME~\cite{flame} parameters that capture dense 3D facial geometry and expression details.
These parameters guide the extraction of high-quality face crops for one-shot or spatio-temporal multi-view layouts, which also serve as visual identity evidence for captioning.
On the audio side, we process the soundtrack separately.
Demucs~\cite{defossez2019demucs} separates speech from background music and effects.
3D-Speaker~\cite{chen20253d} performs speaker diarization to assign speech segments to subjects, and ASR yields per-subject transcripts for later captioning.
SyncNet~\cite{chung2016out} then retains only within-clip appearance-voice pairs whose lip motion is temporally consistent with the speech, rejecting dubbed or off-sync candidates before cross-clip matching and captioning.

\noindent\textbf{Cross-clip Audio-visual Identity Matching.}
To discourage copy-and-paste and enforce non-trivial reference-target gaps, we select identity references from a different clip of the same subject whenever possible.
Clips are first grouped by ArcFace~\cite{deng2019arcface} face embeddings to propose same-person visual candidates.
ERes2Net~\cite{chen2024eres2netv2} then verifies speaker identity and rejects dubbing or voice-over mismatches.
Using the FLAME and transcript annotations from extraction, we retain pairs with varied pose/expression and non-overlapping speech content.
The retained cross-clip pairs provide the appearance and voice references $c^v$ and $c^a$ for their matched target clips while preventing direct content copying.

\noindent\textbf{Multimodal Captioning with Subject Anchors.}
Our multimodal captioning pipeline synthesizes visual, auditory, and textual streams into a structured narrative with precise subject grounding.
To generate text conditions for training, we feed each video clip, its subject-specific face crops, and ASR transcripts to Qwen3-Omni~\cite{qwen-omni} under the subject-anchor prompt in Algorithm~\ref{alg:prompt}.
The model describes camera movement, scene context, ambience and sound effects, as well as each subject's appearance, actions, expressions, and speech.
The prompt binds appearance attributes and spoken content to a unique subject ID, so the resulting structured captions can specify who appears and who speaks in multi-subject scenes.
These captions serve as the text prompts for our joint audio-video generation model.
Algorithm~\ref{alg:prompt} gives the full system prompt used with Qwen3-Omni.

\subsection{Dataset statistics}
As in the main paper, source videos come from OpenHumanVid~\cite{li2025openhumanvid} and OmniHuman~\cite{zhu2026omnihuman}.
We keep clips satisfying $\lvert\mathrm{SyncNet~offset}\rvert{<}3$ and $\mathrm{score}{>}1.5$, together with face similarity ${>}0.72$, speaker similarity ${>}0.70$, and head pose ${<}30^\circ$.
Across the full pipeline these thresholds reject about $18.7\%$ of raw candidates.
The retained training pool covers about $110$k unique identities.
Evaluation uses an identity-disjoint split of $400$ triplets as in the main paper.

\begin{algorithm}[!t]
\caption{System Prompt for Qwen3-Omni}
\label{alg:prompt}
\small
\begin{algorithmic}[1]
\STATE You are a professional video captioning assistant. Generate a fluent, structured natural language description based on the video, audio, speech transcripts, and face images, including the following elements in a logical narrative order.
\STATE \textbf{Camera Movements.} Describe the shot type (e.g., close-up, medium shot), camera movement (e.g., fixed, dolly, pan), angle (e.g., eye level), and position (e.g., front view, over-the-shoulder).
\STATE \textbf{Scene Context.} Describe the setting (e.g., indoor, outdoor), background (e.g., office, red carpet, street), lighting (e.g., soft, high-key, spotlight), and overall atmosphere (e.g., tense, glamorous, nostalgic).
\STATE \textbf{Subject Descriptions (Appearance).} For each main person, use the format: \tagstart{REF\_N}appearance description\tagend{REF\_N}. Include gender, age estimate, hairstyle, facial features, clothing, accessories, and distinguishing traits. Use the provided face images to ensure accurate and consistent identification.
\STATE \textbf{Subject Descriptions (Actions \& Expressions).} Describe body posture, gestures, facial expressions (e.g., serious, smiling), and eye direction.
\STATE \textbf{Subject Descriptions (Speech Content).} For each spoken utterance, identify the speaker via \tagstart{REF\_N}/\tagend{REF\_N}. Use \speechstart{1}spoken text\speechend{1}, \speechstart{2}speech transcript\speechend{2}, etc. Transcribe verbatim. Keep speech synchronized with speaker identity and timing.
\STATE \textbf{Non-Vocal Sounds.} Describe notable background or environmental sounds (e.g., camera shutters, traffic, music, cloth friction), but keep it concise.
\STATE \textbf{Example Output.} This is a classic interview or group photo close-up, shot with a medium-close fixed lens at eye level. The scene is set against a minimalist indoor background, with beige textured wallpaper appearing simple and nostalgic, and soft frontal lighting clearly illuminating the faces. On the left side of the screen, \tagstart{REF\_1}a male with a very short buzz cut and a thin, cool face\tagend{REF\_1} is wearing a dark gray turtleneck sweater. On the right side, \tagstart{REF\_2}a female with short black hair\tagend{REF\_2} is wearing a black top with gold embroidery on the collar, looking gentle. They stand side by side, looking natural. The male looks straight ahead with deep eyes, his expression serious and reserved, while the female tilts her head slightly with a faint smile on her lips. The male says calmly: \speechstart{1}The story has been told.\speechend{1} The female responds gently: \speechstart{2}But the touching part has just begun.\speechend{2} The background sound includes subtle friction of clothes and the quiet sound of air flowing indoors.
\end{algorithmic}
\end{algorithm}

\section{Implementation Details}

Identity-as-Presence is implemented on two dual-tower backbones under the same identity injection mechanism (shared identity embeddings, reference positioning, decoupled asymmetric attention, and subject-anchored captions).
\textbf{Ours-LTX-2.3} is the primary system in the main paper. \textbf{Ours-Ovi}~\cite{low2025ovi} is a reference implementation that demonstrates generality.

Unimodal audio uses $150$k hours of TTS data. Unimodal video uses $100$k identity-labeled clips. Joint multimodal training uses $50$k identity-labeled audio-visual pairs. Stage~3 fine-tuning uses $10$k spatio-temporal multi-view clips.
We use AdamW~\cite{loshchilov2017decoupled} ($\beta_1{=}0.9$, $\beta_2{=}0.999$, $\epsilon{=}10^{-8}$), learning rate $1{\times}10^{-4}$, and batch size $64$.
Unimodal stages optimize only the modality-specific flow-matching loss. Joint training balances towers with $\lambda{=}0.2$.
Reference audio length is $L{=}5$s and $T_{\max}{=}12$s.
Inference uses $50$ steps with audio/video classifier-free guidance of $4.0/3.0$.
Both backbones follow the same three-stage curriculum. Ours-LTX-2.3 uses unimodal audio/video $10$k/$10$k iterations and joint multimodal $15$k steps. Ours-Ovi uses unimodal audio/video $12$k/$8$k iterations and joint multimodal $5$k steps. Both then share the same Stage~3 multi-view fine-tuning.

\section{Evaluation Details}

We group baselines by conditioning paradigm:
\begin{itemize}[leftmargin=*,itemsep=0.2em]
    \item \textbf{R2V}: reference-driven video generation from facial identities (no joint audio synthesis).
    \item \textbf{RA2V}: reference-audio-driven video. Speech is first obtained (e.g., via voice cloning) and then drives lip-synced video.
    \item \textbf{R2AV}: reference-conditioned joint audio-video generation that synthesizes speech and video together from identity cues.
\end{itemize}
Single- and multi-subject splits follow the identity-disjoint benchmark in the main paper. All compared systems use matched identity references and LLM-adapted prompts derived from the same source captions unless otherwise noted.

\subsection{Evaluation settings}
The evaluation set contains $400$ identity-disjoint triplets ($200$ single-subject and $200$ multi-subject).
Each sample provides appearance references (one-shot or spatio-temporal multi-view), a voice reference, and a text prompt with bilingual dialogue.
The single-subject split includes hard cases with large head pose and/or occlusion. The multi-subject split likewise includes hard interaction cases.
For methods that require a reference-based first frame, we use Qwen-Image-Edit~\cite{wu2025qwen} to synthesize an image from the same identity references and adapted prompts.
We observe that OmniCustom~\cite{li2026omnicustom} and ID-LoRA are stronger on English than on Chinese.

\subsection{Evaluation metrics}
\paragraph{Audio metrics.} Following the Seed-TTS protocol~\cite{anastassiou2024seed}, we transcribe generated English and Chinese speech with Whisper-large-v3~\cite{radford2023robust} and Paraformer-zh~\cite{gao2022paraformer}, respectively. We compute the Word Error Rate (WER) between the recognized and target transcripts to measure speech intelligibility.
CLAP~\cite{CLAP} measures audio-text alignment by computing the cosine similarity between the embeddings of the generated audio and the corresponding text prompt.
Fr\'echet Distance (FD) computes distribution distance between generated and real audio.
Audiobox PQ~\cite{audiobox} uses a pretrained no-reference model to predict the human-rated production quality of each audio sample.
AID-SIM computes the cosine similarity between WavLM~\cite{chen2022wavlm} speaker embeddings of the generated and reference speech to assess their speaker identity similarity.

\paragraph{Video metrics.} To assess visual quality, motion dynamics, and prompt alignment, we use three metrics from VBench~\cite{vbench}: Aesthetic Quality (AES), Dynamic Degree (DD), and Overall Consistency (OC).
AES evaluates the aesthetic quality of each generated frame using the LAION~\cite{schuhmann2022laion} aesthetic predictor and averages the resulting scores over the video.
DD estimates optical flow between sampled adjacent frames with RAFT~\cite{RAFT} and reports the proportion of videos classified as dynamic based on a predefined motion threshold.
OC encodes the generated video and its text prompt into a shared embedding space with ViCLIP~\cite{wang2024internvid}, and computes their cosine similarity to measure overall video-text consistency.
To evaluate appearance identity preservation, we additionally use VID-SIM, which averages the cosine similarity between ArcFace~\cite{deng2019arcface} embeddings of generated face crops and their reference faces.

\paragraph{Audio-visual consistency metrics.} To evaluate lip-speech synchronization, we use Sync-C and Sync-D computed by SyncNet~\cite{chung2016out}.
Sync-C measures the confidence of the best temporal alignment between speech and mouth motion, while Sync-D measures the minimum distance between their embeddings across different temporal offsets. Higher Sync-C and lower Sync-D indicate better synchronization.
To evaluate event-level audio-visual alignment, we use ImageBind~\cite{ImageBind}, which computes the cosine similarity between the generated audio and video embeddings.
Face Match (FM) and Speaker Match (SM) for multi-subject binding are defined next.

\paragraph{Face Match and Speaker Match}
Averaged identity similarities (AID-SIM/VID-SIM) are insufficient for multi-subject evaluation:
a model can still score well when both identities appear, yet are \emph{cross-bound} or partially collapsed.
Motivated by recurring dual-subject failure modes, we report two assignment-based metrics on multi-subject prompts only (the $200$ multi-subject cases). Single-subject clips are excluded because unique assignment among $K{>}1$ references is undefined.

Given $K$ reference faces and $K$ reference voices, we extract face tracks (detection + tracking) and speech segments (VAD + diarization).
Each track/segment is embedded with ArcFace~\cite{deng2019arcface} and WavLM~\cite{chen2022wavlm}.
Bipartite matching to references with empirical similarity thresholds yields face and voice assignments $\pi_v$ and $\pi_a$.

\noindent\textbf{Face Match (FM).}
FM is the fraction of reference faces that are \emph{uniquely} matched to a generated track above threshold.
It penalizes face reuse, missing faces, and wrong visual assignment.

\noindent\textbf{Speaker Match (SM).}
A reference voice~$k$ counts as matched only if all of the following hold.
(i)~It is uniquely assigned a speech support above the speaker-similarity threshold.
(ii)~\emph{Turn exclusivity:} that support overlaps speech assigned to any other matched speaker by at most $\tau$ seconds (we use $\tau{=}0.3$). Otherwise both overlapping speakers are unmatched.
(iii)~\emph{Face-voice consistency:} within the temporal support of each of its matched segments, the face track with the highest SyncNet~\cite{chung2016out} lip-sync confidence must be assigned to the same identity~$k$ under $\pi_v$.
Segments without a reliable active face count as failures for~(iii).
SM is the fraction of reference voices that satisfy (i)--(iii).
Thus SM penalizes missing speakers and voice collapse, dual talking under expected turn-taking, and appearance-voice swap even when unimodal coverage looks complete.

A high FM with low SM indicates speaker-side or cross-modal binding failures (including swap and dual talking). Simultaneous drops typically reflect identity collapse or missing subjects.
Main-paper tables report FM and SM under this protocol.

\begin{figure}[t]
    \centering
    \suppincludegraphics[width=0.99\columnwidth]{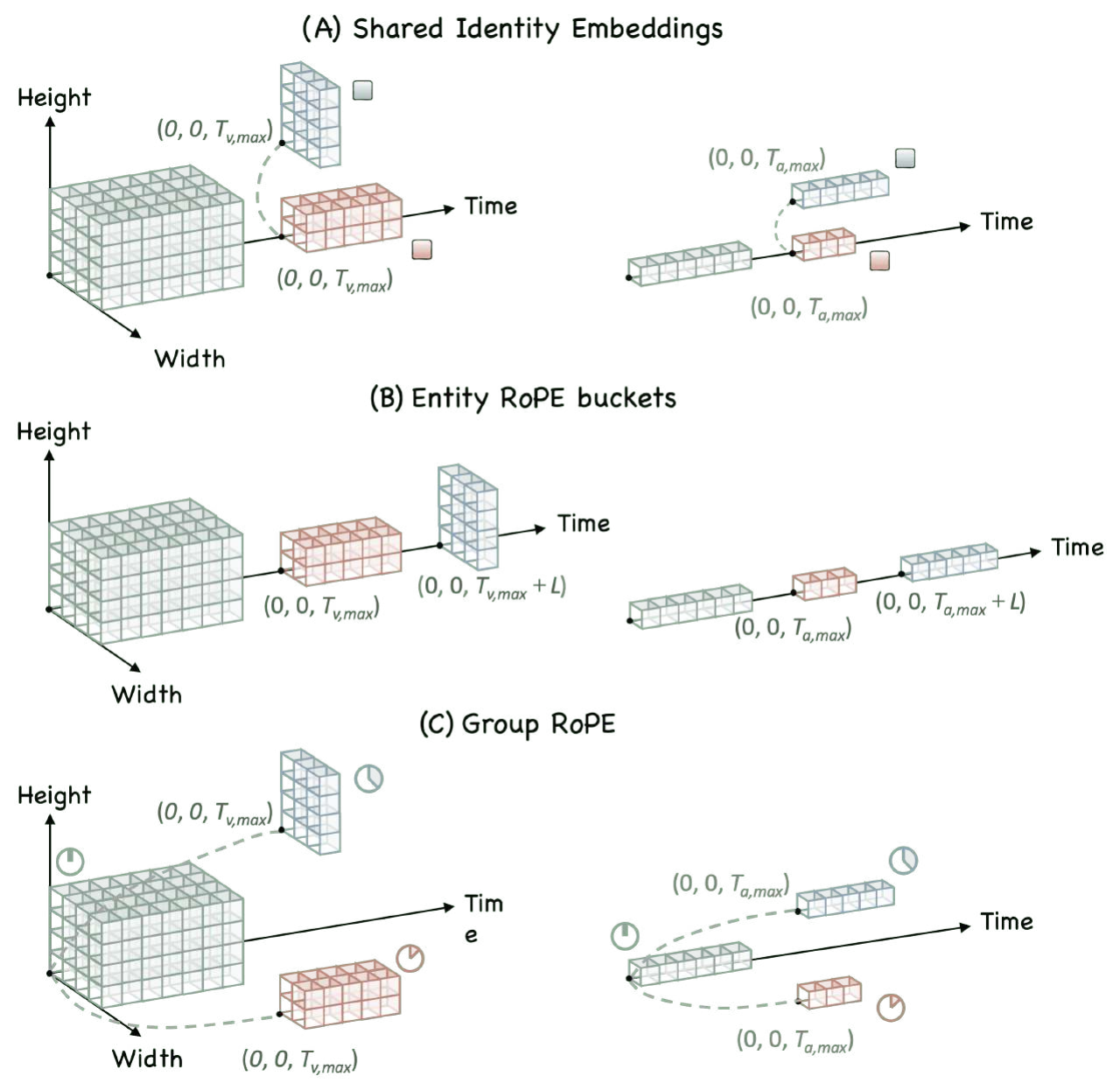}
    \caption{Three appearance-voice binding variants aligned with the main-paper ablation.
    (A)~Shared Identity Embeddings (IE) assign the same learnable embedding to each subject's visual and audio references.
    (B)~RoPE Offset (Entity RoPE buckets) places different subjects in separate temporal RoPE ranges.
    (C)~Rotary Bias (Group RoPE) retains shared temporal positions and distinguishes subjects with group-specific rotary biases.
    Green denotes noisy target tokens. Red and blue denote references from two subjects.}
    \label{fig:supp_binding_variants}
\end{figure}

\section{Additional Ablation Details}

Figure~\ref{fig:supp_binding_variants} illustrates three alternatives for encoding subject identity in visual and audio reference tokens.
All variants use the same LTX-2.3 dual-tower backbone and differ only in the binding signal.
\begin{itemize}[leftmargin=*,itemsep=0.2em]
    \item \textbf{A~Shared Identity Embeddings (IE):} all subjects share the same RoPE start positions, and the visual and audio reference tokens of each subject~$k$ receive the same learnable identity embedding $e^{id}_{(k)}$.
    \item \textbf{B~RoPE Offset (Entity RoPE buckets):} following Syn-RoPE-style positioning~\cite{guodreamid}, we remove IE and assign each subject a non-overlapping temporal RoPE range beyond the generation horizon.
    \item \textbf{C~Rotary Bias (Group RoPE):} references are placed at $T_{v,\max}{=}0$ and $T_{a,\max}{=}0$ (not beyond the generation horizon as in A), and a group rotary bias $R_{\mathrm{group}}(g)$ aligns same-person cross-modal references.
\end{itemize}
A is most reliable on FM/SM.
Unlike B, it does not entangle identity with the temporal axis or rely on large RoPE offsets as the subject count grows.
C reuses in-timeline reference indices with a continuous-phase prior, which is less explicit than a shared discrete identity code.
Text-anchored RoPE, such as SA-MRoPE~\cite{chen2026omni}, is similarly geometric and complements the caption structure rather than replacing A.

Subject anchors act on the text side rather than on reference RoPE.
Dense captions often mix speakers and blur who should appear. Anchored spans tie appearance and speech turns to the same subject ID, which is why A+SA improves multi-subject grounding beyond token-level IE alone.

On the training side, the main-paper curriculum table already shows one-stage joint training lagging two-stage and full multi-view fine-tuning under matched budgets.
The last stage mainly adapts the model to the two complementary reference layouts. Spatial packing vs.\ temporal expansion under motion is shown qualitatively in Section~``Additional Qualitative Results''.

\section{Additional Qualitative Results}

We expand the qualitative analysis in the main paper along cross-domain generalization, text-reference disentanglement, and spatio-temporal multi-view conditioning, and provide additional single-/multi-subject results including non-vocal sound.

\subsection{Cross-domain generalization.}
\begin{figure*}[!t]
    \centering
    \suppincludegraphics[width=0.99\textwidth]{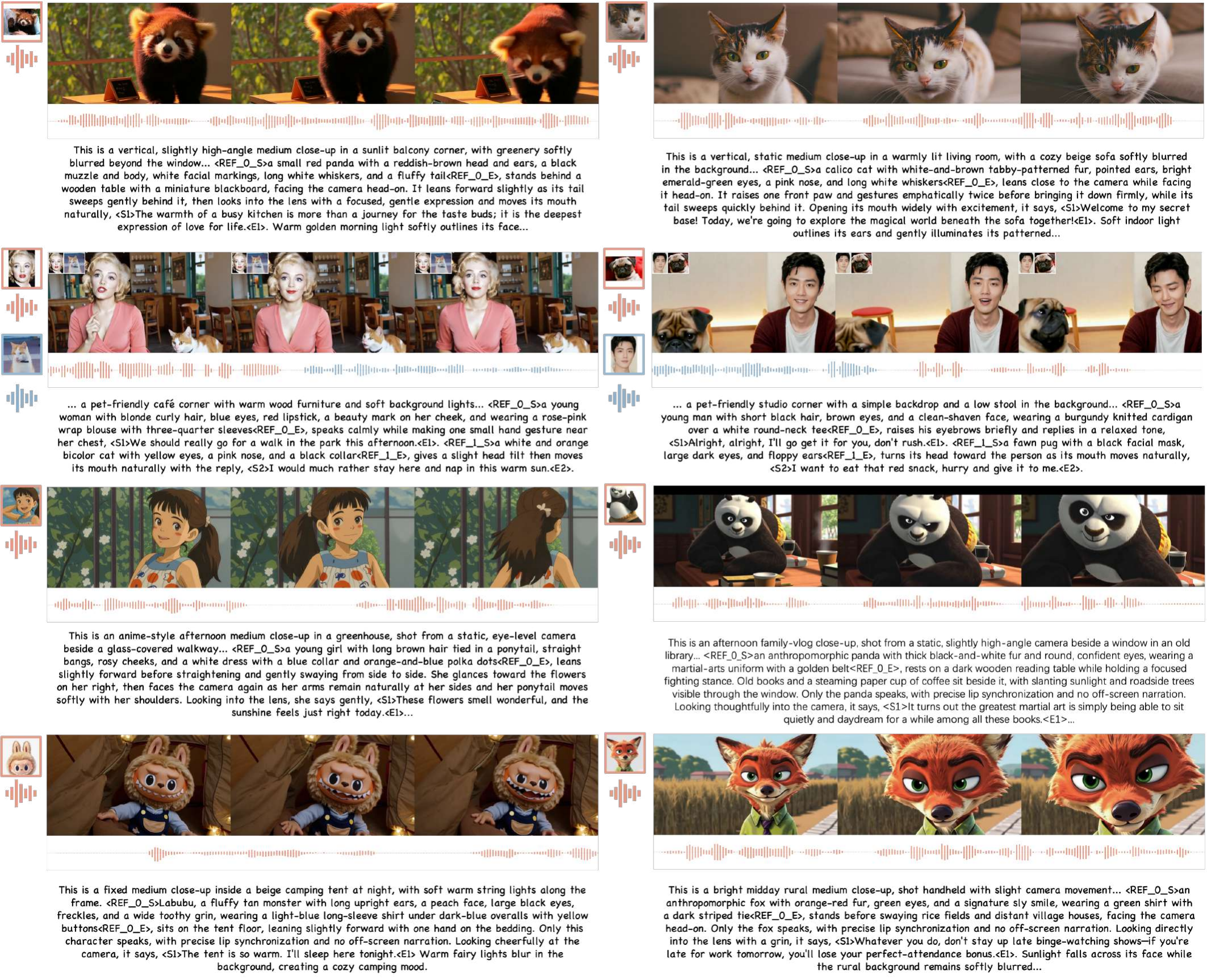}
    \caption{Cross-domain personalization.}
    \label{fig:supp_case_ood}
\end{figure*}

Although training uses only real-human identities, Figure~\ref{fig:supp_case_ood} shows that our model can personalize cartoon and animal subjects given matching visual/voice references.

\subsection{Text-reference disentanglement.}
\begin{figure*}[!t]
    \centering
    \suppincludegraphics[width=0.96\textwidth]{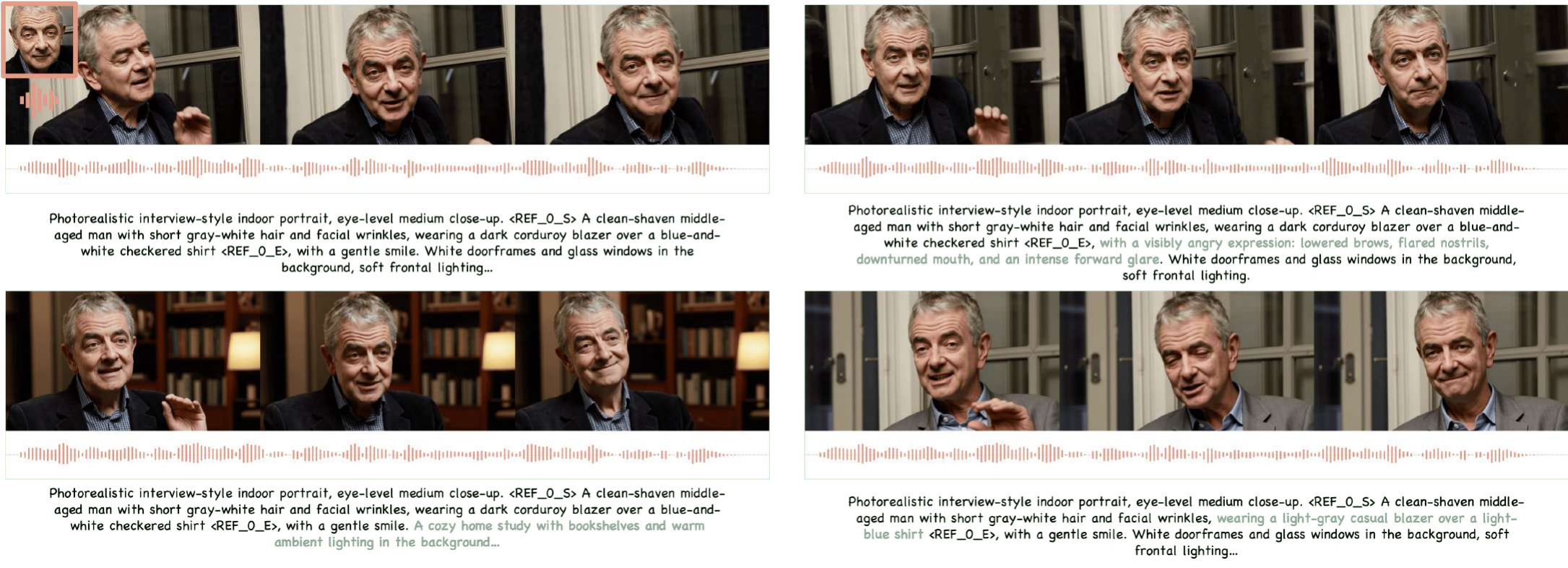}
    \caption{Fixed reference, prompt edits (L$\rightarrow$R): init generation; change expression; change scene; change scene+outfit.}
    \label{fig:supp_case_disentangle}
\end{figure*}

Figure~\ref{fig:supp_case_disentangle} starts from an \emph{init} generation that matches the reference ID, scene, and outfit, then edits the prompt left to right: \emph{change expression}, \emph{change scene}, and \emph{change outfit}.
The reference anchors identity. Text controls which attributes are rewritten and which are maintained.

\subsection{Spatiotemporal multi-view consistency.}
\begin{figure*}[!ht]
    \centering
    \suppincludegraphics[width=0.99\textwidth]{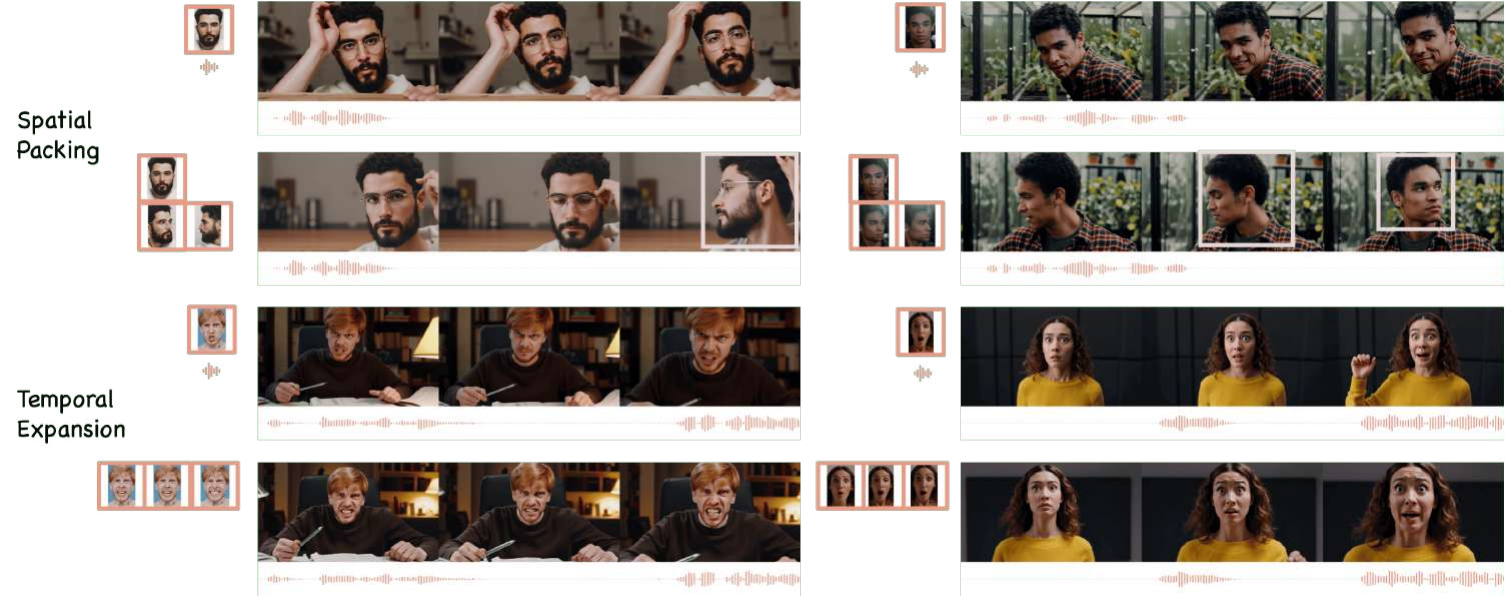}
    \caption{Spatiotemporal multi-view references.}
    \label{fig:supp_case_multiview}
\end{figure*}

Figure~\ref{fig:supp_case_multiview} compares one-shot injection with spatial packing and temporal expansion.
A single reference image provides limited identity evidence. It lacks multi-view appearance and short-term dynamics such as expression change.
Spatial packing aggregates complementary viewpoints for view-robust identity, while temporal expansion injects ordered frames that capture dynamic appearance cues over time.

\subsection{Single-subject personalized generation}
\begin{figure*}[!ht]
    \centering
    \suppincludegraphics[width=0.96\textwidth]{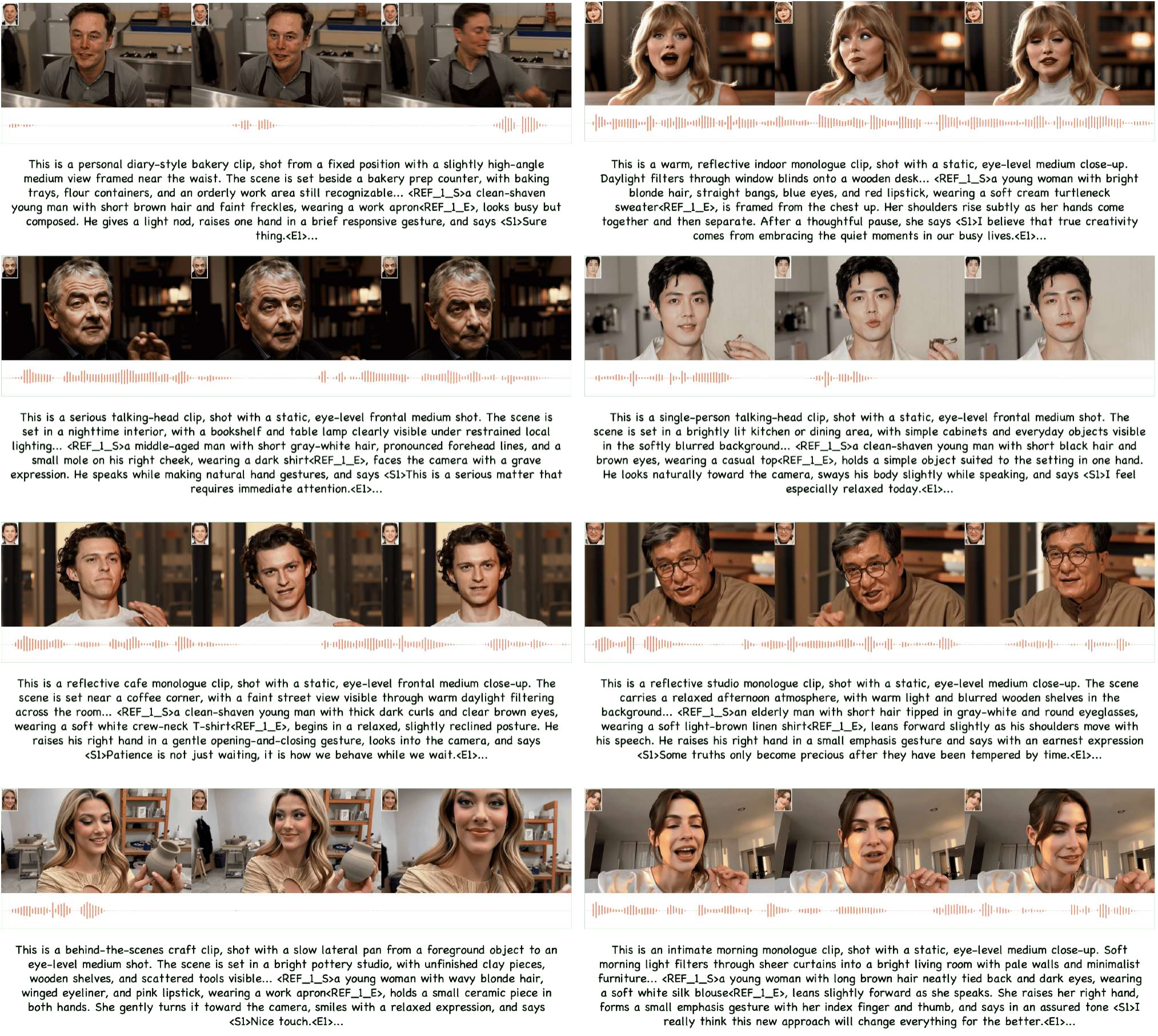}
    \caption{Qualitative results of single-subject personalized joint audio-video generation.}
    \label{fig:supp_single}
\end{figure*}

\begin{figure*}[!ht]
    \centering
    \suppincludegraphics[width=0.99\textwidth]{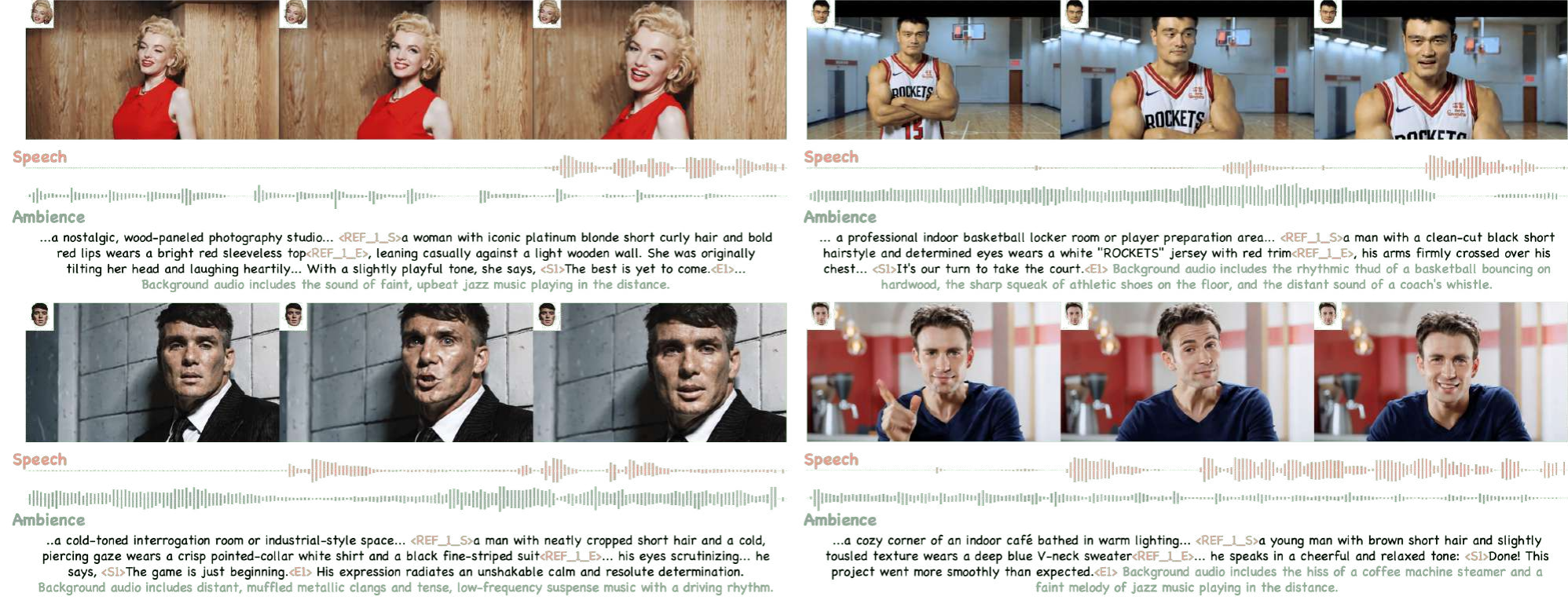}
    \caption{Responding to audio descriptions in the prompt, the model jointly synthesizes identity-preserving speech and ambient sounds that stay consistent with the video.}
    \label{fig:supp_sep_audio}
\end{figure*}

Figure~\ref{fig:supp_single} presents additional results for single-subject personalized joint audio-video generation. These results verify that our model consistently preserves both facial appearance and vocal timbre across diverse prompts and contexts.
Figure~\ref{fig:supp_sep_audio} further shows that the model follows audio descriptions in the prompt and generates speech together with environmental sounds that remain harmonious with the video.

\subsection{Multi-subject personalized generation}
\begin{figure*}[!ht]
    \centering
    \suppincludegraphics[width=0.99\textwidth]{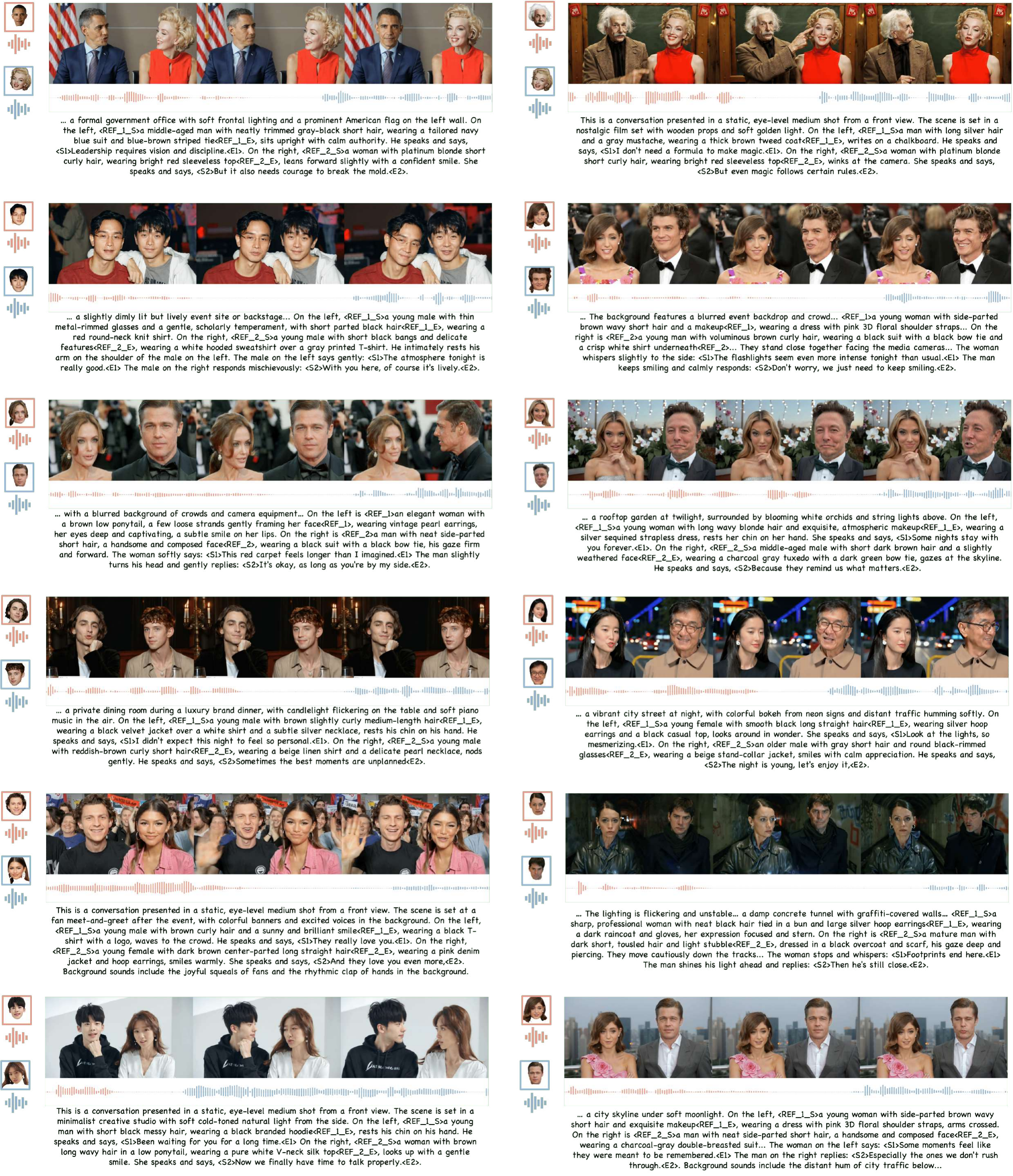}
    \caption{Qualitative results of multi-subject personalized joint audio-video generation.}
    \label{fig:supp_multi}
\end{figure*}

We further evaluate multi-subject scenarios. As shown in Figure~\ref{fig:supp_multi}, our approach preserves the unique characteristics of each subject while maintaining high visual fidelity.

\section{Limitations}

High-quality identity-labeled audio-visual pairs remain scarce relative to unimodal corpora.
The multi-stage curriculum mitigates modality imbalance but cannot invent missing joint diversity, so rare accents, crowded multi-party dialogue, and atypical cameras stay underrepresented.
Filtering for lip-sync, pose, and speaker purity further biases training toward clean studio-like clips.
End-to-end R2AV prioritizes appearance-voice binding and audio-visual coherence, so we do not claim to match cascade systems whose speech comes from oracle TTS on ground-truth transcripts.
Finally, training uses short real-human clips from $4$ to $10$ seconds.
Minute-scale or streaming generation is not validated.
Even though Figure~\ref{fig:supp_case_ood} shows encouraging cross-domain generalization, cartoon and animal results remain zero-shot relative to the real-human training distribution.

\begin{figure*}[!ht]
    \centering
    \suppincludegraphics[width=0.99\textwidth]{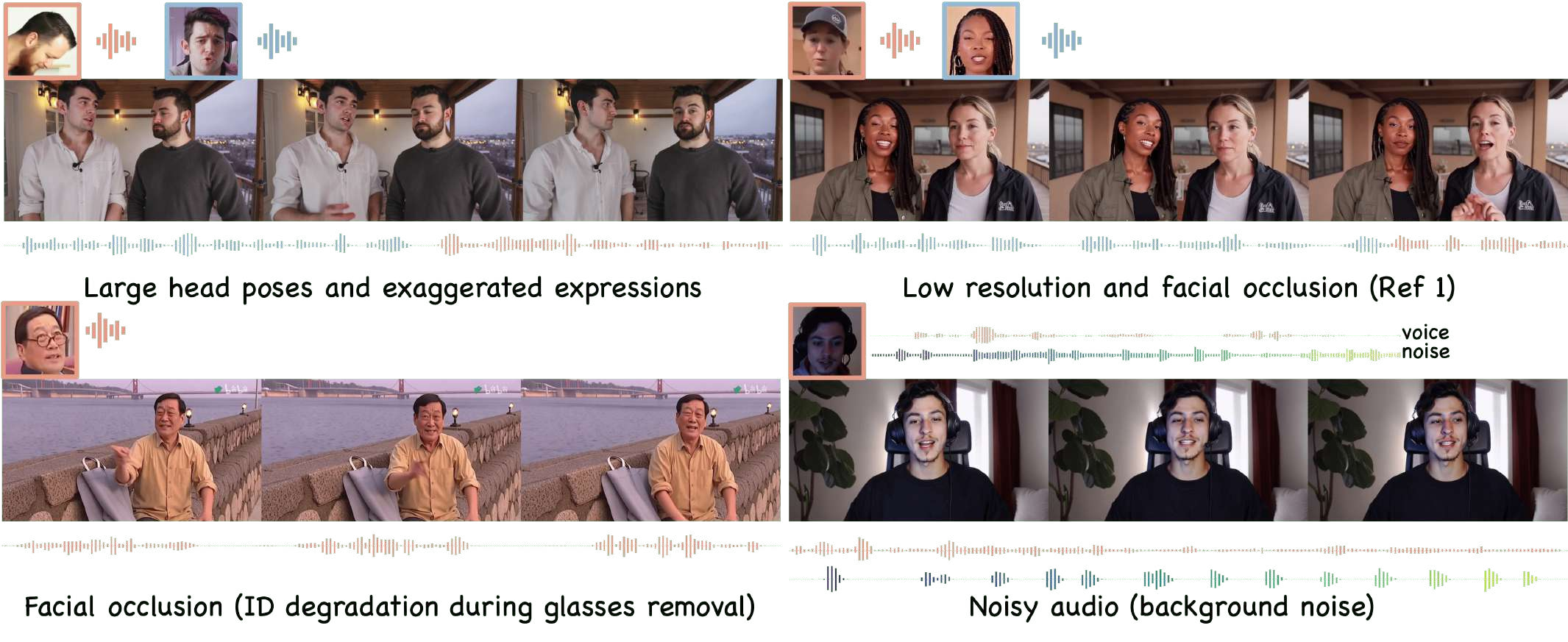}
    \caption{Failure cases: large head pose, heavy facial occlusion, extreme expression/pose, and noisy reference audio.}
    \label{fig:failure}
\end{figure*}

These biases surface when references leave that clean regime (Figure~\ref{fig:failure}). Large head pose and heavy occlusion thin identity cues, extreme expression or motion can break lip-expression consistency, and noisy or cross-talk voice references weaken timbre cloning, with intermittent speaker collapse in multi-subject clips.
Spatio-temporal multi-view references help on hard pose but do not remove these cases.


\end{document}